%% file: ifacconf.tex
\documentclass{ifacconf}
\usepackage{graphicx}      
\usepackage{natbib}        
\usepackage{comment}
\usepackage{tabularx}
\usepackage{booktabs}
\usepackage{subcaption}
\usepackage{caption} 
\usepackage{mathtools}
\usepackage{amsmath}    
\usepackage{amssymb}    
\usepackage{graphicx}   
\usepackage{bm}         
\usepackage{url}
\usepackage{graphicx}      

\usepackage{xcolor}
\usepackage{float}      
\usepackage{placeins}
\usepackage{multirow}
\usepackage{bbm}
\usepackage{extdash}
\usepackage{array}     
\usepackage{boldline}  
\usepackage{siunitx}
\usepackage{natbib}
\usepackage{lipsum}

\begin{document}
\begin{frontmatter}

\title{Component-Aware Pruning Framework for Neural Network Controllers via Gradient-Based Importance Estimation}





\thanks[footnoteinfo]{Authors have contributed equally.}


\author[First]{Ganesh Sundaram} 
\author[First]{Jonas Ulmen} 
\author[First]{Daniel Görges}


\address[First]{Department of Electrical and Computer Engineering, \\ RPTU University Kaiserslautern-Landau, Germany \\ 
    (e-mail: \{ganesh.sundaram, jonas.ulmen, daniel.goerges\}@rptu.de).}

\begin{abstract}                
The transition from monolithic to multi-component neural architectures in advanced neural network controllers poses substantial challenges due to the high computational complexity of the latter. Conventional model compression techniques for complexity reduction, such as structured pruning based on norm-based metrics to estimate the relative importance of distinct parameter groups, often fail to capture functional significance. This paper introduces a component-aware pruning framework that utilizes gradient information to compute three distinct importance metrics during training: Gradient Accumulation, Fisher Information, and Bayesian Uncertainty. Experimental results with an autoencoder and a TD-MPC agent demonstrate that the proposed framework reveals critical structural dependencies and dynamic shifts in importance that static heuristics often miss, supporting more informed compression decisions.
\end{abstract}

\begin{keyword}
AI-driven modeling and control, Machine learning for modeling and prediction, Reinforcement learning and deep learning in control,
Neural network Compression, Structured pruning, Neural network controllers
\end{keyword}

\end{frontmatter}

\section{Introduction}
\input{introduction}

\section{Related Work}
\label{sec:related_work}
\input{related_works/related_work_haiku}

\section{Methodology}
\label{sec:methodology}

This section presents our primary contributions. First, we propose three importance metrics for each pruning group, each derived from the gradients of the corresponding parameters. Second, we introduce a training regime that accumulates gradients using an exponential moving average and incorporates a scheduling mechanism to prevent vanishing gradients in the importance metric as the model converges on the task loss.

\subsection{Gradient-Based Importance Metrics}
\label{subsec:grad_based_metrics}

Although frameworks such as \texttt{Torch-Pruning} offer essential tools for managing architectural dependencies during structured pruning, they do not specify methods for measuring the importance of each parameter group~\citep{Fang_2023_CVPR}. Users are therefore required to define appropriate scoring metrics to guide pruning decisions. Traditional approaches often use static, heuristic-based metrics, such as weight norms. However, these can be inadequate for complex models in which a group's functional importance does not necessarily align with its magnitude. To address this limitation, this work investigates dynamic, gradient-based metrics that quantify a group's influence on task loss, thereby providing a more principled basis for pruning decisions.

Previous approaches, such as~\citep{sundaram2025application}, employ grid search or gradient-based optimization to determine pruning-group coefficients, which specify the number of parameters each group must remove to achieve the desired pruning objective. However, these coefficient weightings can only be determined post-training, limiting their utility for importance estimation during training. To address this limitation, a set of gradient-based importance metrics is introduced, computed online, and integrated into the training process. These metrics leverage group-wise gradients of the training loss and capture complementary aspects of importance, including instantaneous sensitivity, curvature, and accumulated learning activity. Combining these metrics provides a more robust estimate of importance than relying on a single metric.

\subsubsection{Gradient Magnitude Accumulation}
\label{subsub:gradient}

For a pruning group \(g\) with parameter set \(\Theta_g\), define the per-parameter average absolute gradient at one iteration \(t\) of an epoch as
\[
I_g^{\mathrm{grad}}(t)
\;=\;
\frac{1}{N_g}\sum_{\theta \in \Theta_g}\left|\nabla_\theta\mathcal{L}(t)\right|,
\]
Here, $\nabla_\theta \mathcal{L}$ represents the gradient of the loss $\mathcal{L}$ with respect to the parameters $\theta$, and $N_g \in \mathbb{N}$ denotes the number of parameters in the group $g$. This metric quantifies the model's immediate tendency to update a group, reflecting the steepness of the loss landscape for that group and the group's instantaneous importance. Higher values indicate groups that are updated more frequently during training. While this metric captures instantaneous learning activity, it may be subject to noise.

\subsubsection{Fisher Information (Diagonal Approximation)}
\label{subsub:fisher}

The Fisher Information Matrix quantifies the sensitivity of the loss to parameter changes. Parameters with high Fisher information have a strong influence on model predictions. It is computed as the diagonal approximation of the Fisher Information Matrix, i.e.

$$I_g^{\text{Fisher}}(t) = \frac{1}{N_g} \sum_{\theta \in \Theta_g} \mathbb{E}[(\nabla_\theta \mathcal{L}(t))^2]$$

In practice, we approximate the expectation with the current batch, i.e.

$$I_g^{\text{Fisher}}(t) \approx \frac{1}{N_g} \sum_{\theta \in \Theta_g} (\nabla_\theta \mathcal{L}(t))^2$$

Fisher information quantifies the curvature of the loss landscape and indicates the sensitivity of the loss function to small parameter perturbations. Parameters associated with large Fisher values exert a strong influence on model predictions and are thus more critical for overall performance. Additionally, the Fisher matrix provides a tractable approximation for the Hessian, as it is the expected outer product of gradients and is formulated as

$$\mathcal{F}(t) = \mathbb{E}[\nabla \mathcal{L}(t) \cdot \nabla \mathcal{L}(t)^T] \approx \mathbb{E}[\nabla^2 \mathcal{L}(t)]$$

\subsubsection{Bayesian (Empirical)}
\label{subsub:bayesian}

In this context, standard Bayesian inference is challenging because it requires specifying a likelihood for the gradient observations and computing posterior integrals at each training step. Both requirements are often computationally expensive or intractable. Therefore, a lightweight empirical Bayes formulation is adopted.

The per-group gradient energies are modeled as samples from an exponential distribution with the rate parameter $\lambda_g$, i.e.
$E_g \sim \mathrm{Exponential}(\lambda_g)$ with density
\[
p(E_g \mid \lambda_g) = \lambda_g e^{-\lambda_g E_g}, \; E_g \ge 0.
\]
The conjugate prior for the rate $\lambda_g$ is a Gamma distribution $\Gamma(\alpha_g,\beta_g)$, which enables simple additive updates. Instead of maintaining the full posterior $p(\lambda_g \mid \text{data})$, only its mean is tracked via the sufficient statistics $(\alpha_g,\beta_g)$. To capture cumulative evidence of a group's learning activity, the per-iteration group gradient energy is defined as follows.
\[
E_g^{(t)} \;=\; \frac{1}{N_g}\sum_{\theta \in N_g}\left|\nabla_\theta \mathcal{L}(t)\right|.
\]
We maintain a Gamma prior in the $(\text{shape}=\alpha,\ \text{rate}=\beta)$ parameterization and update it online with a fractional pseudo-count $\kappa>0$:
\[
\alpha_g^{(t+1)} \;=\; \alpha_g^{(t)} + \kappa, \qquad
\beta_g^{(t+1)} \;=\; \beta_g^{(t)} + \kappa\cdot\frac{E_g^{(t)}}{\eta},
\]
Here, $\eta>0$ is a scaling factor and $\kappa$ controls the effective observation weight (with $\kappa=0.25$ used in experiments). The posterior mean of the rate is then
\[
\mu_g^{(t)} \;=\; \frac{\alpha_g^{(t)}}{\beta_g^{(t)}},
\]
This value summarizes the accumulated learning activity for group $g$. Subsequently, this historical activity is combined with the current sensitivity as follows.
\[
I_g^{\mathrm{Bayes}}(t)
\;=\;
\log\!\bigl(1+\mu_g^{(t)}\bigr)\,\bigl(1 + I_g^{\mathrm{Fisher}}(t)\bigr),
\]
where the $\log(1+\cdot)$ term improves numerical stability as $\mu_g$ grows and modulates instantaneous Fisher sensitivity with long-term activity.

\textit{Interpretation:} The Gamma model interprets large, persistent gradient energies as evidence that the network actively utilizes a group. The Gamma prior accumulates this evidence over time, and its mean $\mu_g^{(t)}$ provides a smoothed estimate of long-term group activity. Consequently, the Bayesian importance score $I_g^{\mathrm{Bayes}}$ assigns higher ranks to groups that are both currently sensitive to the loss (exhibiting high Fisher information) and historically active (with large $\mu_g^{(t)}$), thereby prioritizing groups that consistently contribute rather than those relevant only during isolated updates.

\subsection{Training Regime}
\label{subsec:implementation_remarks}

Training the network for subsequent pruning with the proposed metrics requires additional gradient information, which must be collected during training. Two phenomena need to be addressed. First, gradient information can be noisy when the model has not yet converged on the task loss. To mitigate this, we employ temporal smoothing. Second, as the model approaches convergence, gradients become small, leading to group importance scores that are numerically indistinguishable. To address this, we apply cosine scheduling, which serves as a controlled perturbation to the importance regularization factor $\lambda$.

\subsubsection{Temporal Smoothing}
\label{subsub:temporal_smoothing}

Exponential smoothing using the exponential moving average (EMA) is applied to each raw importance signal $I_g(t)$ to reduce batch-to-batch variance
\[
\tilde{I}_g(t) = \gamma\,\tilde{I}_g(t-1) + (1-\gamma)\,I_g(t),
\]
where the smoothing coefficient $\gamma$ is selected from the interval $[0,1)$, with a common default value of $\gamma=0.9$. The resulting smoothed score $\tilde{I}_g$ is then used for ranking and pruning decisions.

\subsubsection{Coefficient Parameters}
\label{subsub:parameters}
The pseudo-count $\kappa$, the scale $\eta$, and the EMA coefficient $\gamma$ are tunable parameters. The fractional update $\kappa$ facilitates the gradual accumulation of evidence and reduces abrupt changes during the early iterations. If necessary, the three signals (gradient magnitude, Fisher information, and Bayesian estimates) can also be combined using a weighted sum or a multiplicative approach to rank groups by their smoothed importance $\tilde{I}_g$ when selecting pruning candidates.

\subsubsection{Loss Function Formulation}
\label{subsub:loss_function}

The total loss function is formulated as
\[
    \mathcal{L}_{\text{total}}(t) = \mathcal{L}_{\text{task}}(t) + \lambda_{\text{weight}}(t) \mathcal{L}_{\ell_1}(t),
\]
where the $\ell_1$ regularization term is defined as
\[
    \mathcal{L}_{\ell_1}(t) = \sum_{i=1}^{G} \lambda_i(t) \|\Theta_i\|_1.
\]
In this formulation, $\mathcal{L}_{\text{task}}$ denotes the task-specific loss, such as the mean squared error for reconstruction tasks. The parameter $\lambda_{\text{weight}}$ is a global hyperparameter that controls the $\ell_1$ penalty. The term $\lambda_i(t)$ represents the time-varying regularization strength for group $i$, while $\|\Theta_i\|_1$ denotes the $\ell_1$ norm of the parameters in group $i$. The number $G \in \mathbb{N}$ denotes the number of groups.

In the initial stages of training, the relatively small value of $\lambda_{\text{weight}}$ ensures that $\mathcal{L}_{\text{task}}$ is the dominant component of the total loss. As a result, the optimizer primarily minimizes the task-specific loss. After the task loss converges, the influence of the $\ell_1$ regularization term increases, directing the optimizer to reduce the magnitude of the weight parameters and encourage sparsity.

\subsubsection{Scheduler for Regularization Coefficients}
\label{subsub:cyclic_scheduler}

To determine which pruning groups contribute most significantly to the total loss under $\ell_1$ regularization, a per-group cosine schedule with phase offsets is applied to the regularization coefficients $\lambda_i(t)$ during training. The coefficient for group $i$ at epoch $t$ is defined as
\[
    \lambda_i(t) = \frac{1}{\sqrt{N_{i}}} \, S(t,i),
\]
where
\[
    S(t,i) = \lambda_{\min} + \bigl(\lambda_{\max} - \lambda_{\min}\bigr)
    \cdot \frac{1}{2}\left[1 + \cos\!\left(2\pi \frac{t + \phi_i}{T}\right)\right],
\]
and the phase offset for group $i$ is $\phi_i = \frac{i}{n} \, T$.

In this context, $\lambda_{\text{base}}$ represents the base regularization strength (default: $10^{-5}$). The minimum regularization level is $\lambda_{\min} = 0.1 \lambda_{\text{base}}$, and the maximum is $\lambda_{\max} = 2.0 \lambda_{\text{base}}$. The parameter $T$ specifies the cycle length in epochs (default: $T = 20$). The factor $\sqrt{N_i}$ normalizes the coefficient by the number of parameters in group $i$ to ensure a fair comparison across groups.

During training, different groups attain the maximum value of $\lambda_i(t)$ at various times, resulting in differential exposure to sparsity pressure. Groups that sustain or increase their weights during periods of intense regularization are identified as structurally critical. In contrast, groups whose weights consistently decline under strong regularization are regarded as less critical and are thus more suitable for pruning.

\begin{table*}[!tbp]
\centering
\setlength{\tabcolsep}{8pt}
\renewcommand{\arraystretch}{0.99}
\caption{Model characteristics and training specifications of Autoencoder and TD-MPC}
\label{tab:model_parameters}
\begin{tabular}{@{}lc lc lc@{}}
\toprule
\multicolumn{2}{c}{\textbf{Autoencoder}} &
\multicolumn{4}{c}{\textbf{TD-MPC}} \\
\cmidrule(lr){1-2} \cmidrule(lr){3-6} 
\textbf{Parameter} & \textbf{Value} 
& \textbf{Parameter} & \textbf{Value}
& \textbf{Parameter} & \textbf{Value} \\
\midrule
Input Size      & $28\times28$ pixels   & Input Size        & $28\times28$ pixels  & Total FLOPs & $4.12 \times 10^{8}$  \\
Latent Dim.     & 8/64/256/512                   & Latent Dim.       & 50                  &  Hidden Units   &  512\\
Components      & Encoder, Decoder      & Components        & Encoder, Dynamics, Value, Policy, Reward  &    &               \\
Parameters      & 2,312,992            & Total Parameters  & 1,548,360         &                &                 \\
Layers/Comp.    & 2                    & Layers/Comp.  & 3                    &                &                 \\
Model Size      & \SI{8.84}{MB}        & Model Size    & \SI{12.4}{MB}        &                &                 \\
Peak PSNR       & $22.34\,\mathrm{dB}$ & Episode Reward & 858.6                &                &                 \\
Final MSE       & 0.002238             & Episode Length & 125 steps          &                &                 \\
\bottomrule
\bottomrule
\end{tabular}
\end{table*}

\section{Experimental Setup}
\label{sec:experimental_step}

To evaluate the proposed methodology, two complementary use cases are presented.

\subsection{Use Cases}
\label{subsec:usecases}

The first model is a simple autoencoder that, despite its simple architecture, incorporates latent-space encoding—a fundamental component of advanced neural network architectures—and enables the visualization of the consequential effects of pruning on reconstruction performance. The second model is the state-of-the-art model-based reinforcement learning framework, TD-MPC, which also uses latent representations to examine their influence on neural network controllers (NNCs). Table~\ref{tab:model_parameters} summarizes the characteristics and training details for both the autoencoder and TD-MPC.

\subsubsection{Autoencoder on MNIST}

The autoencoder serves as a canonical test model, trained to compress MNIST dataset inputs into low-dimensional latent representations using an \emph{encoder} and to reconstruct the original images from these compressed codes using a \emph{decoder}. The MNIST dataset, a widely used benchmark, contains grayscale images of handwritten digits and is commonly employed to evaluate image models and representation learning capabilities~\citep{zhu2018classification}. The architecture learns compact features that preserve essential visual content, and the model is trained end-to-end to minimize reconstruction loss.

\subsubsection{TD-MPC for Balancing an Inverted Pendulum}

TD-MPC is a model-based reinforcement learning framework that combines model predictive control (MPC) with a learned latent state representation~\citep{hansen2022temporaldifferencelearningmodel}. This framework trains a latent-dynamics model and a terminal-value function via temporal-difference learning, and conducts short-horizon planning directly in the latent space. The TD-MPC architecture comprises five neural components: an encoder \(h_{\theta}\) that maps pixel observations \(s_k\) to latent states \(z_k\), a dynamics model \(f_{\theta}\) that predicts the next latent state \(z_{k+1}\) given \(z_k\) along with the action \(a_k\), a reward model \(r_{\theta}\) that estimates immediate rewards, a value function \(V_{\phi}\) that predicts long-term returns, and a policy \(\pi_{\psi}(a \mid z)\) trained to imitate the MPC planner for rapid deployment.

The proposed method is evaluated on the inverted pendulum swing-up task from OpenAI Gym~\citep{brockman2016openai}, where the agent receives only raw pixel observations. The encoder compresses these images into a low-dimensional latent space, and the learned world model predicts future transitions within this space. This latent representation enables efficient MPC-based planning for both the swing-up maneuver and the subsequent pendulum stabilization.

\subsection{Component-Aware Pruning Group Formation}
\label{subsec:comp_aware_groups_formation}

Our previous work extends dependency-graph-based pruning to component-aware pruning for MCNAs by forming component-specific groups~\citep{sundaram2025enhancedpruningstrategymulticomponent}. Instead of generating a single monolithic graph, this method constructs separate dependency graphs for each component. This decomposition enables the identification of both component-specific pruning groups and inter-component coupling groups. As a result, pruning decisions become more flexible by allowing the selective removal of groups while preserving essential architectural and functional elements. Table~\ref{tab:pruning_groups_combined} provides a detailed breakdown of the groups formed, including constituent modules and parameter counts for both trained use case models.

\begin{table*}[ht]
\centering
\setlength{\tabcolsep}{4.5pt}
\renewcommand{\arraystretch}{0.95}
\sisetup{group-separator={,}}
\caption{Pruning group decomposition for MNIST Autoencoder and TD-MPC showing component-specific and coupling groups with constituent modules and parameter counts.}
\label{tab:pruning_groups_combined}
\begin{tabular}{@{}lcccccccc@{}}
\toprule
 & & & \multicolumn{2}{c}{\textbf{Autoencoder}} & \multicolumn{4}{c}{\textbf{TD-MPC}} \\
\cmidrule(r){4-5}\cmidrule(l){6-9}
\textbf{Group type} & \textbf{Component} & \textbf{Group} &
\textbf{Modules} & \textbf{Parameters} &
\textbf{Component} & \textbf{Group} & \textbf{Modules} & \textbf{Parameters} \\
\midrule
\multirow{7}{*}{\textit{Comp.-specific}} 
  & Encoder & 1 & 8 & \num{202148}    & Encoder    & 1 & 1 & \num{25632} \\
  &         & 2 & 4 & \num{339560}    & Encoder    & 2 & 1 & \num{9248}  \\
  & Decoder & 1 & 8 & \num{339692}    & Encoder    & 3 & 1 & \num{9248}  \\
  &         & 2 & 4 & \num{202280}    & Dynamics   & 1 & 1 & \num{262656} \\
  &         &   &   &                 & Reward     & 1 & 1 & \num{262656} \\
  &         &   &   &                 & Pi         & 1 & 1 & \num{262656} \\
  &         &   &   &                 & Q1         & 1 & 1 & \num{262656} \\
  &         &   &   &                 & Q2         & 1 & 1 & \num{262656} \\
\midrule
\multirow{2}{*}{\textit{Coupling}}
  & Encoder-Decoder & 1 & 1 & \num{99584}   & Encoder-Pi                & 1 & 2 & \num{26112}   \\
  &                 &   &   &               & EncoderPi-Dyn.Rew.Q1Q2    & 1 & 5 & \num{369152}  \\
\bottomrule
\bottomrule
\end{tabular}
\end{table*}

\subsection{Estimating Pruning Group Importance}
\label{subsec:imp_groups}

\subsubsection{Limitations of Conventional Approaches} 
\label{subsubsec:limitations_imp_estimation}

Structured pruning involves making binary decisions to retain or remove entire groups of weights. Conventional approaches typically use norm-based metrics, such as the $\ell_1$ or $\ell_2$ norm, to assess group importance. In these methods, groups are ranked by their norms, and the lowest-ranked groups are eliminated to achieve the desired compression ratio. Although this strategy may be effective for general-purpose models, it often fails to address the needs of specialized, application-specific networks. A primary limitation is that a low norm does not always correspond to low importance, as such groups may still contain information essential for model stability, robustness, or performance. As a result, reliance on norm-based scoring can lead to substantial performance loss, underscoring a significant limitation of conventional approaches for estimating the functional importance of pruning groups.

\subsubsection{Component-Specific and Coupling Groups} 
\label{subsubsec:mcna_groups}

An MCNA, as described in Section~\ref{subsec:comp_aware_groups_formation}, consists of both component-specific and coupling groups. Previous studies have frequently relied on heuristics or educated guesses to determine which groups to prune, leading to several common hypotheses about their relative importance. The adoption of gradient-based importance estimation offers a more systematic and principled alternative, eliminating the need for such assumptions. This data-driven method enables the systematic evaluation of these established hypotheses, providing deeper insights into the model's architecture and clarifying the contributions of different groups to overall performance.

The primary hypotheses that have guided earlier pruning strategies are as follows:

\begin{hypo} (Coupling):
Coupling groups are considered the most critical elements, regardless of their size. These groups function as informational bottlenecks, transmitting essential data for component integration, and are thus typically excluded from pruning.
\end{hypo}

\begin{hypo} (Early Layers): 
Groups linked to components located at the initial stages of the network architecture are essential for maintaining system functionality. Consequently, these groups are generally retained during pruning.
\end{hypo}

\begin{hypo} (Static Importance):
The significance of each group is assumed to be static, remaining constant throughout the training process and unchanged after training concludes.
\end{hypo}

\section{Results and Discussion}
\label{sec:results_discussion}

To evaluate the validity of the proposed importance recommendations and their implications towards the proposed hypotheses, a canonical MCNA architecture use case, the autoencoder, was examined. Latent dimension sizes $d \in \{8, 64, 256, 512\}$ were systematically varied to represent conditions ranging from a highly constrained information bottleneck to an over-parameterized regime. The methodology was also assessed in a control application (Use Case 2). In both scenarios, importance scores were computed using Gradient Accumulation, Fisher Information, and Bayesian uncertainty metrics.

\subsection{Use Case 1: MNIST Autoencoder}

\begin{figure*}[t!]
    \centering
    \begin{subfigure}{\textwidth}
        \centering
        \includegraphics[width=0.32\linewidth]{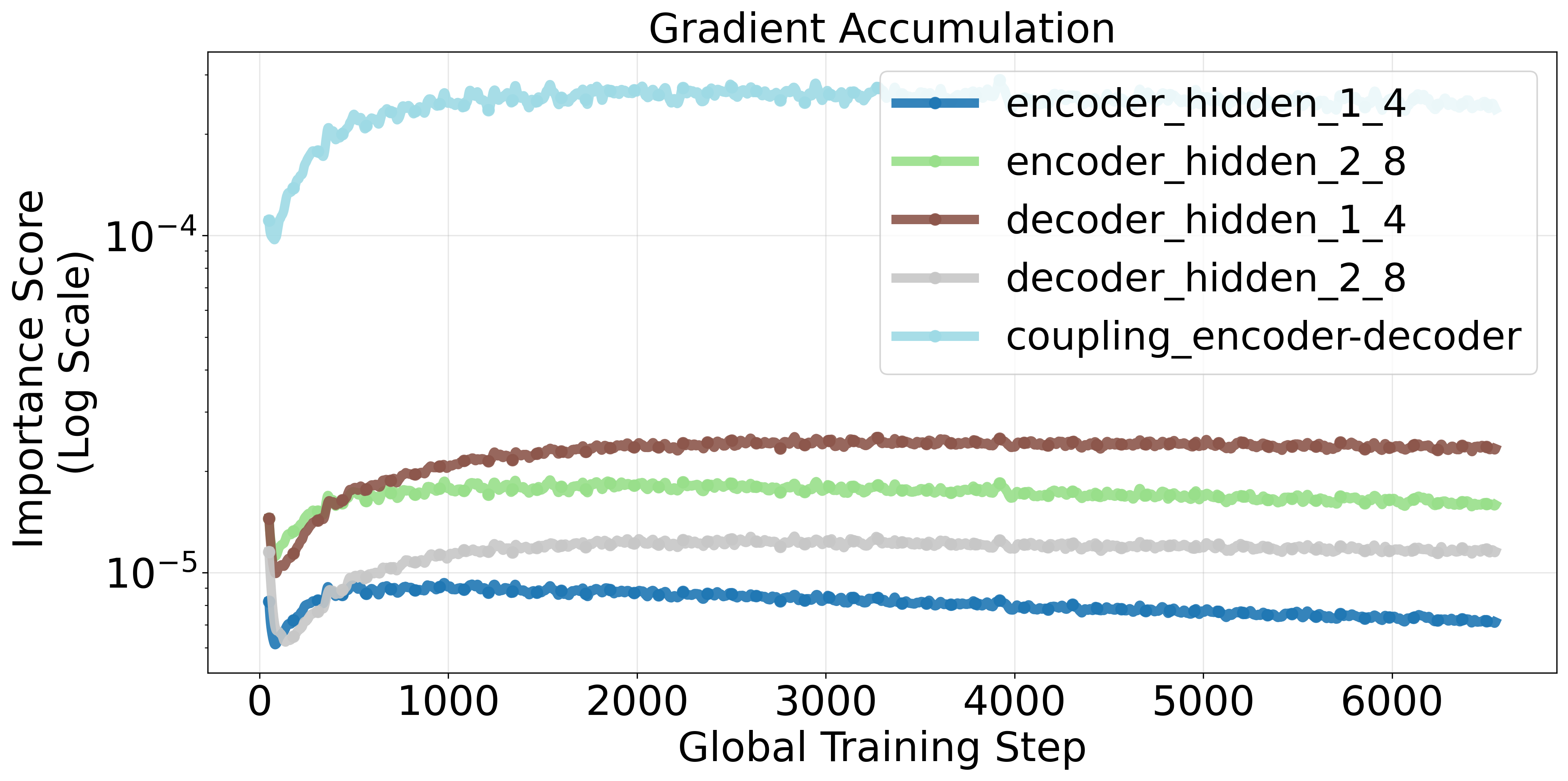} \hfill
        \includegraphics[width=0.32\linewidth]{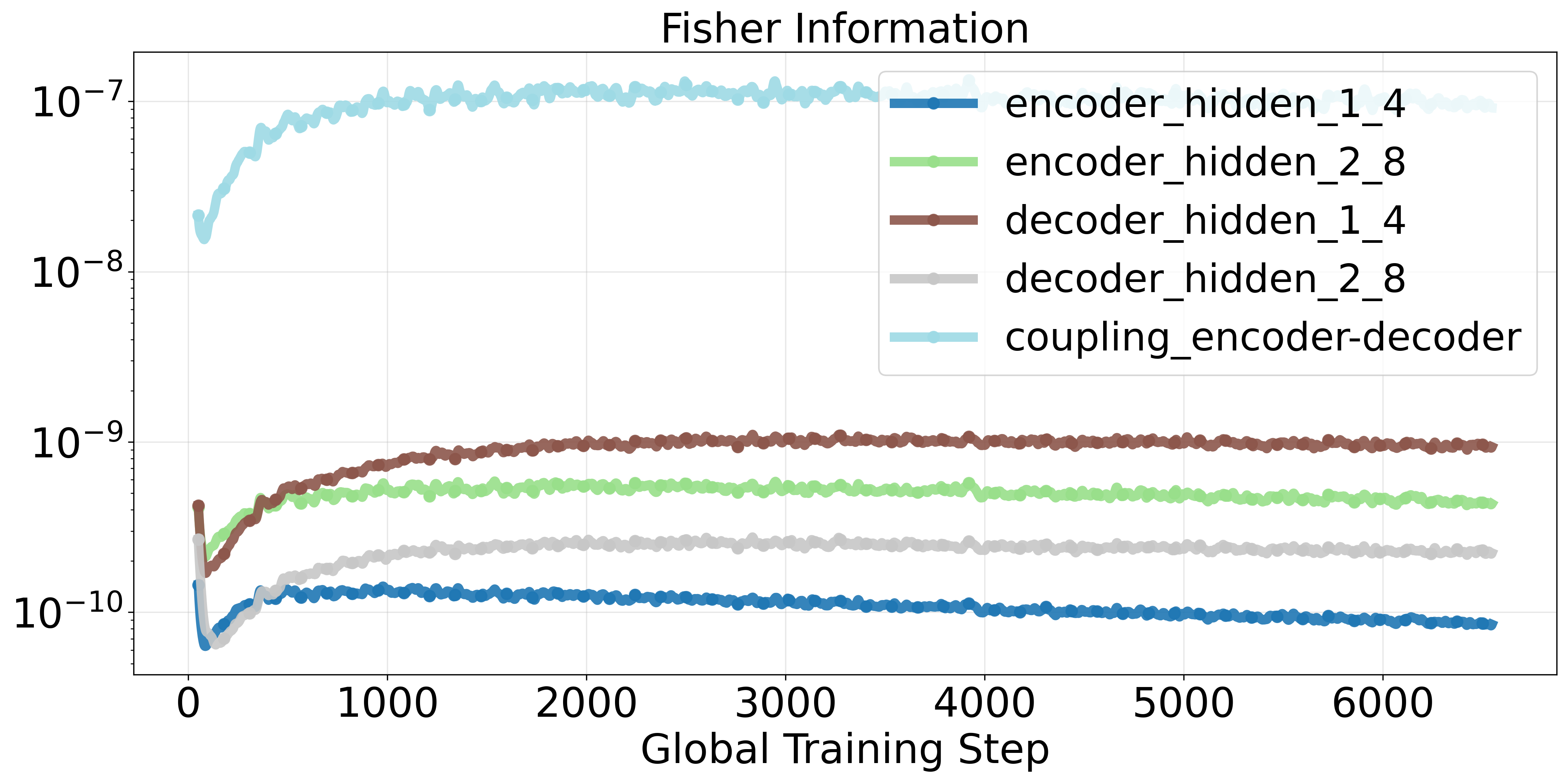} \hfill
        \includegraphics[width=0.32\linewidth]{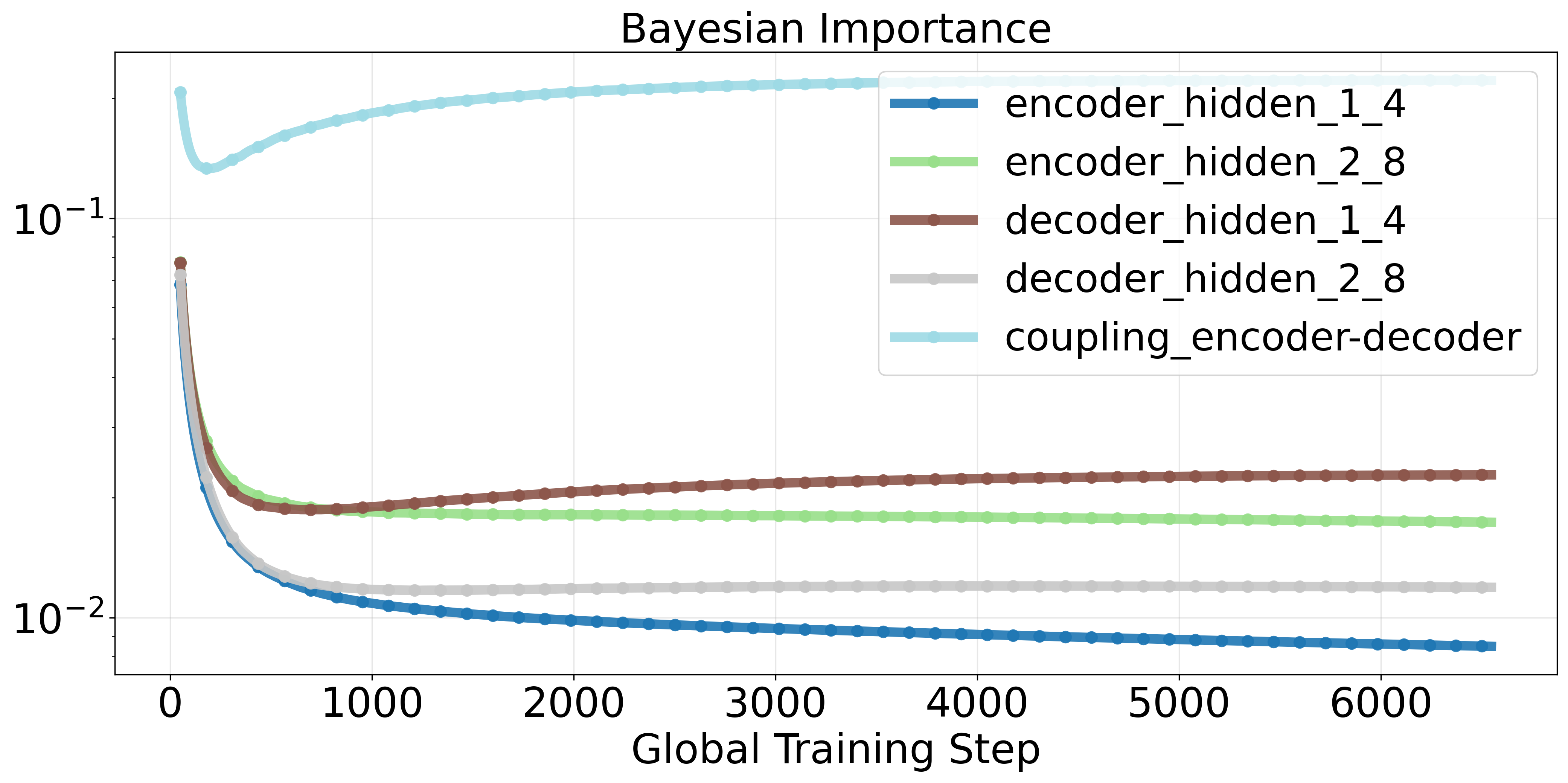}
        \caption{Latent Dimension 8 (Highly Constrained)}
        \label{fig:row1}
    \end{subfigure}
    \vspace{0.2cm}

    \begin{subfigure}{\textwidth}
        \centering
        \includegraphics[width=0.32\linewidth]{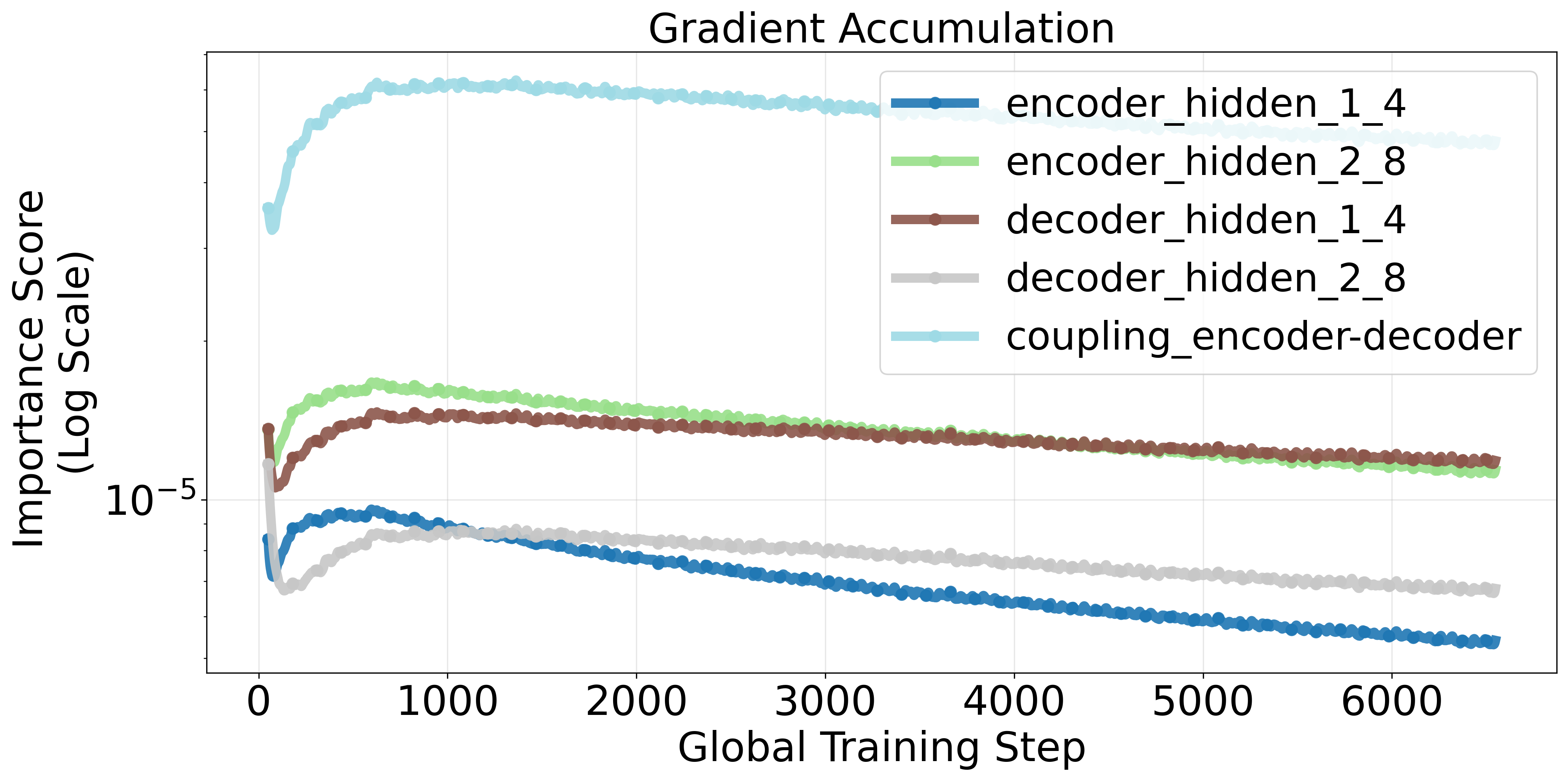} \hfill
        \includegraphics[width=0.32\linewidth]{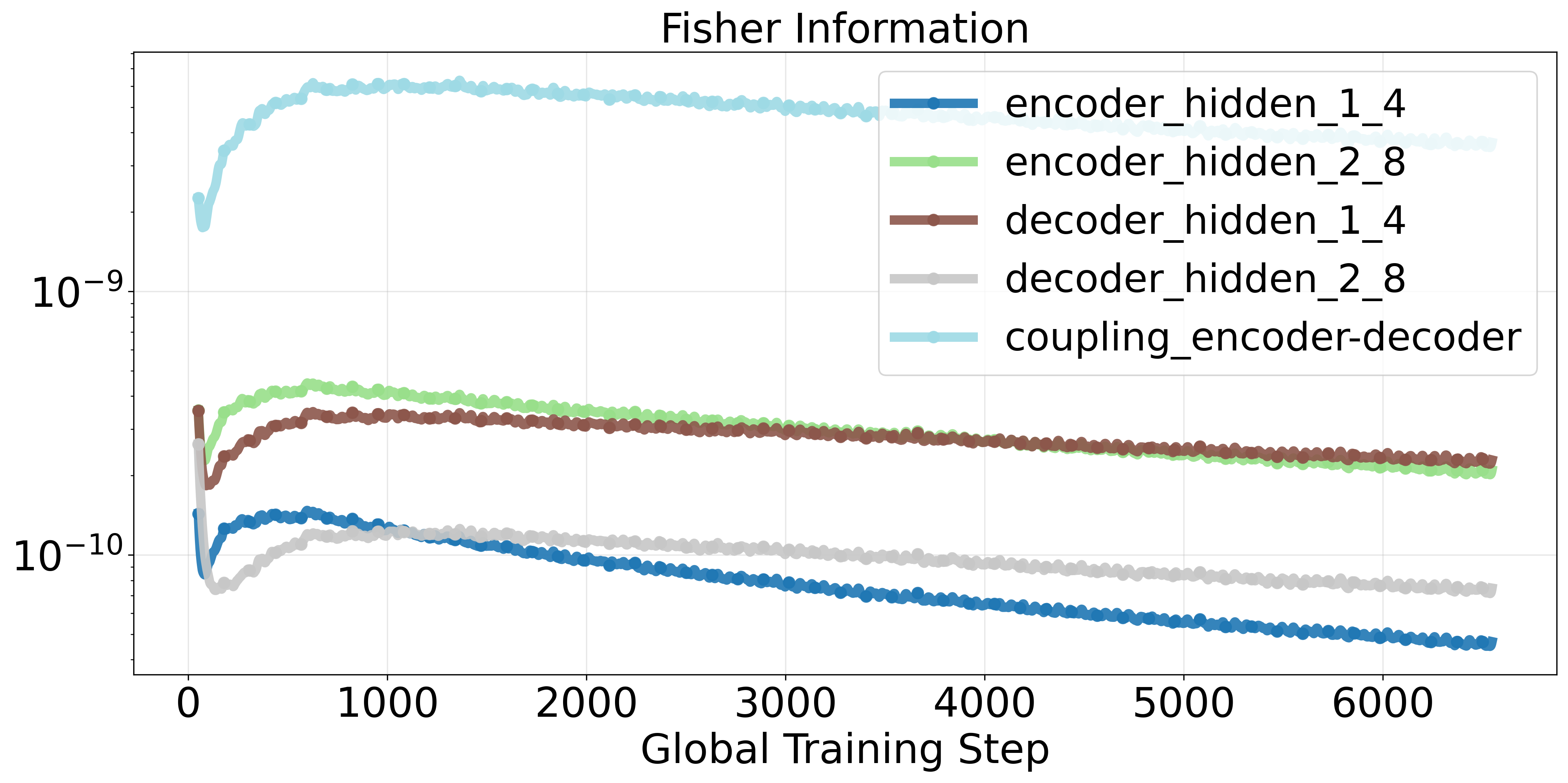} \hfill
        \includegraphics[width=0.32\linewidth]{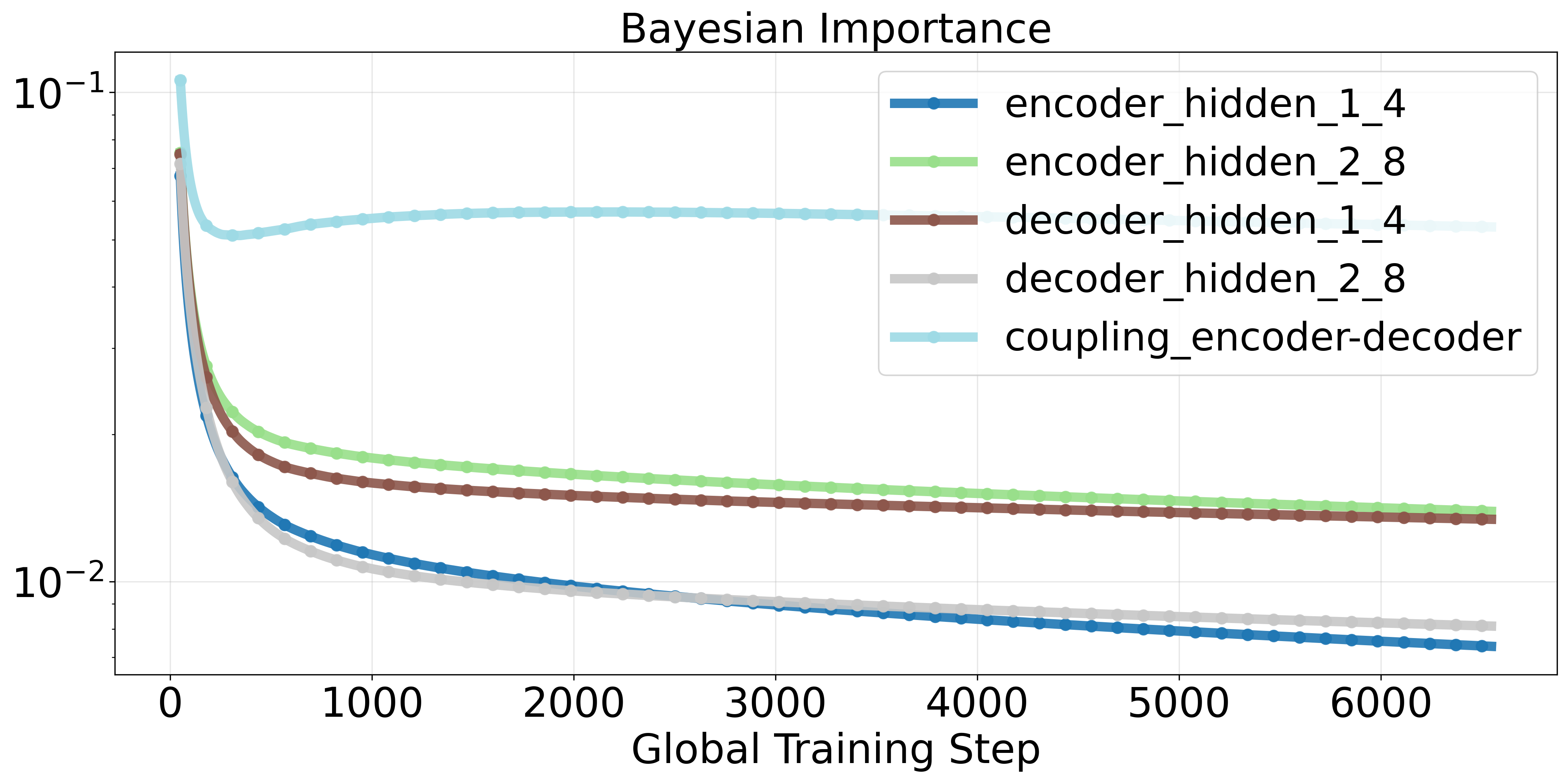}
        \caption{Latent Dimension 64}
        \label{fig:row2}
    \end{subfigure}
    \vspace{0.2cm}

    \begin{subfigure}{\textwidth}
        \centering
        \includegraphics[width=0.32\linewidth]{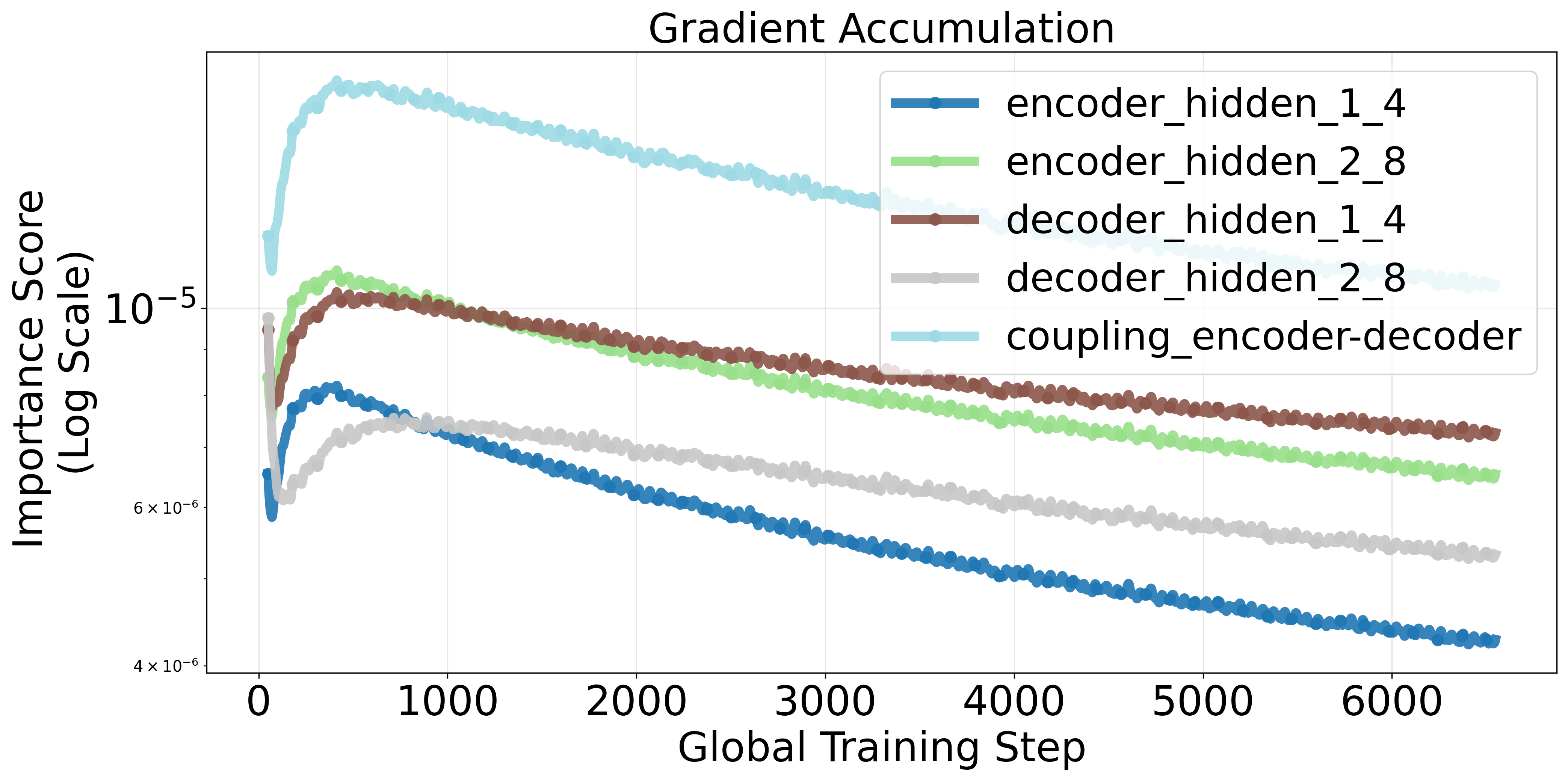} \hfill
        \includegraphics[width=0.32\linewidth]{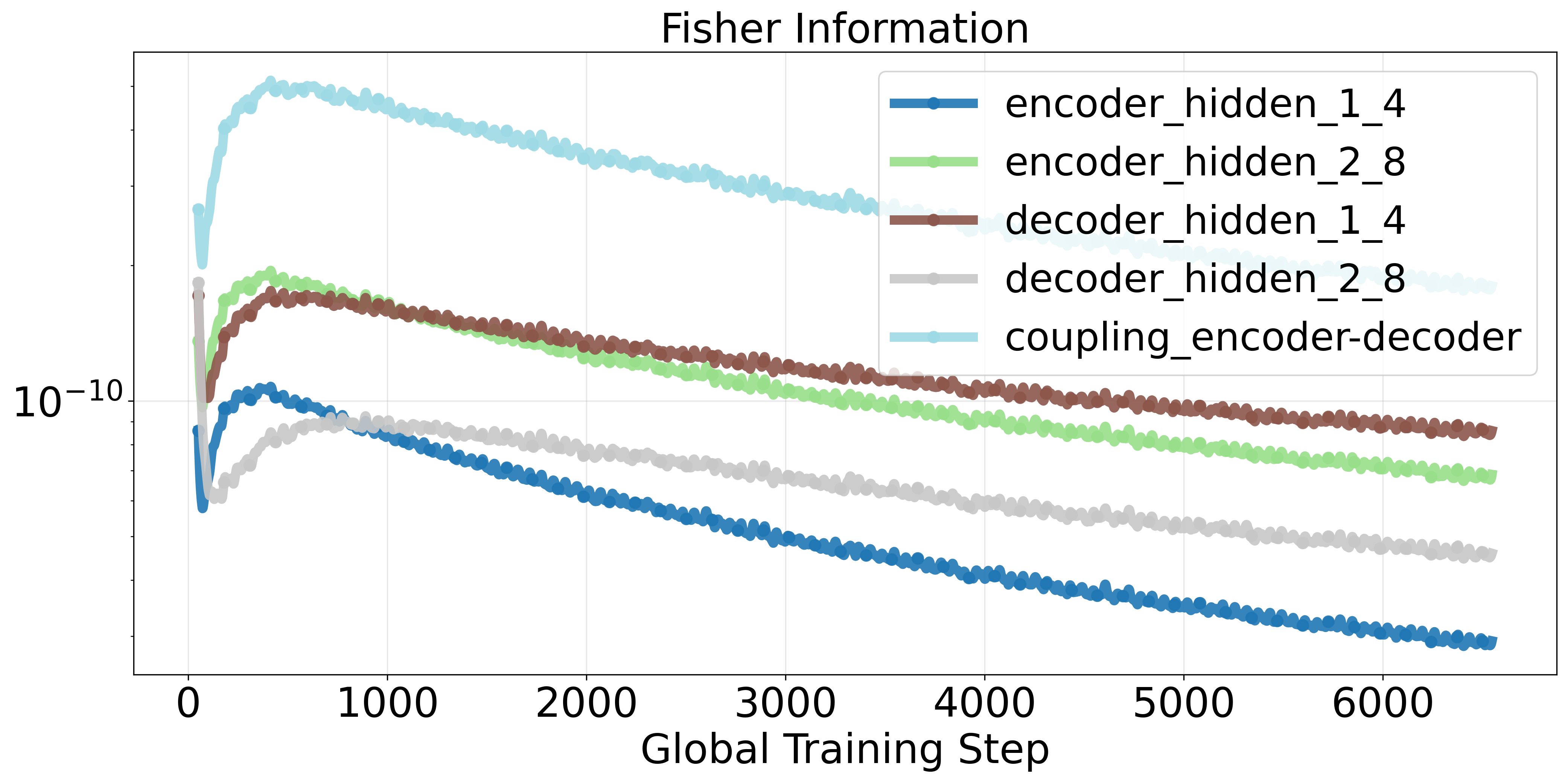} \hfill
        \includegraphics[width=0.32\linewidth]{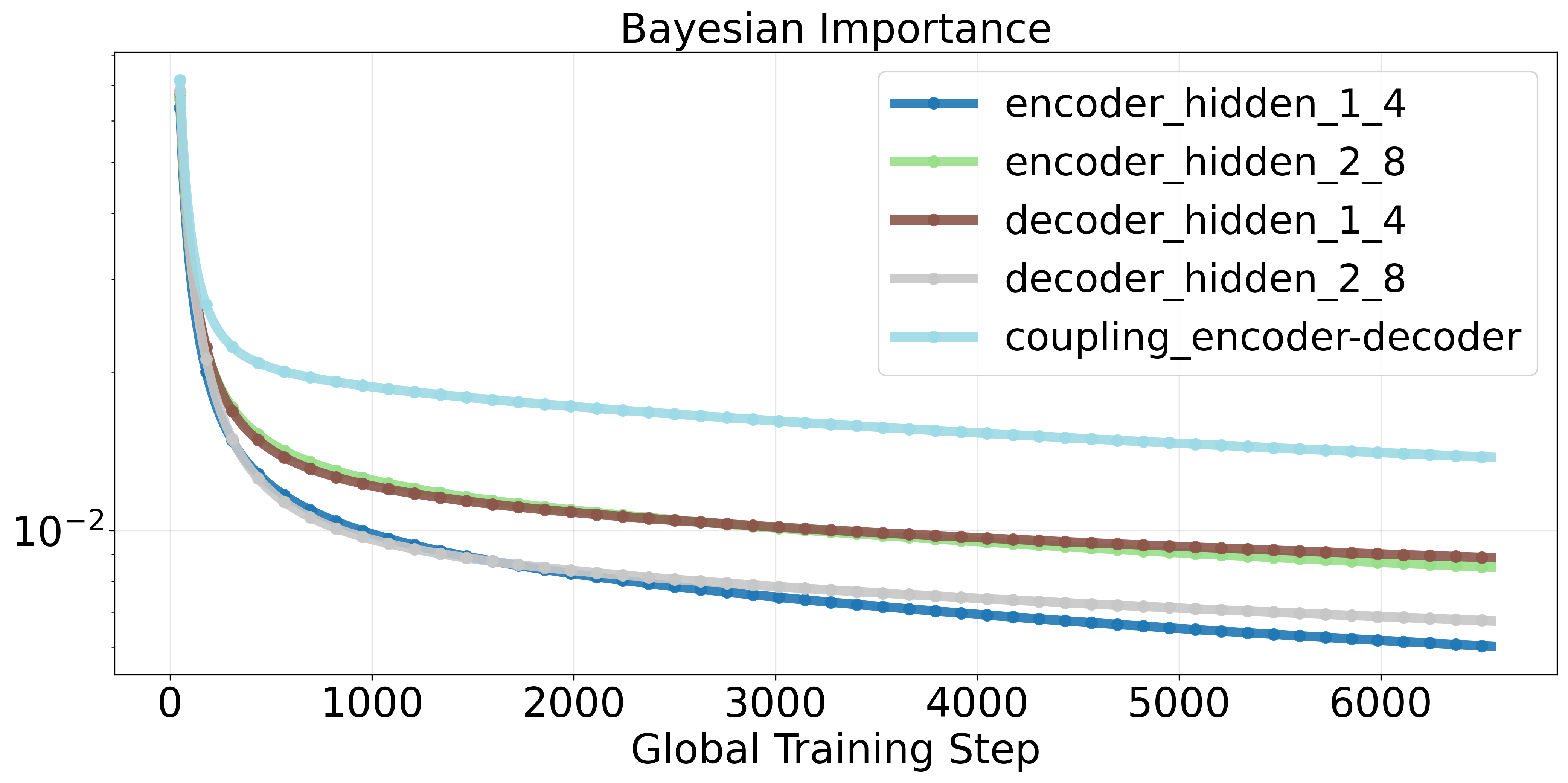}
        \caption{Latent Dimension 256}
        \label{fig:row3}
    \end{subfigure}
    \vspace{0.2cm}

    \begin{subfigure}{\textwidth}
        \centering
        \includegraphics[width=0.32\linewidth]{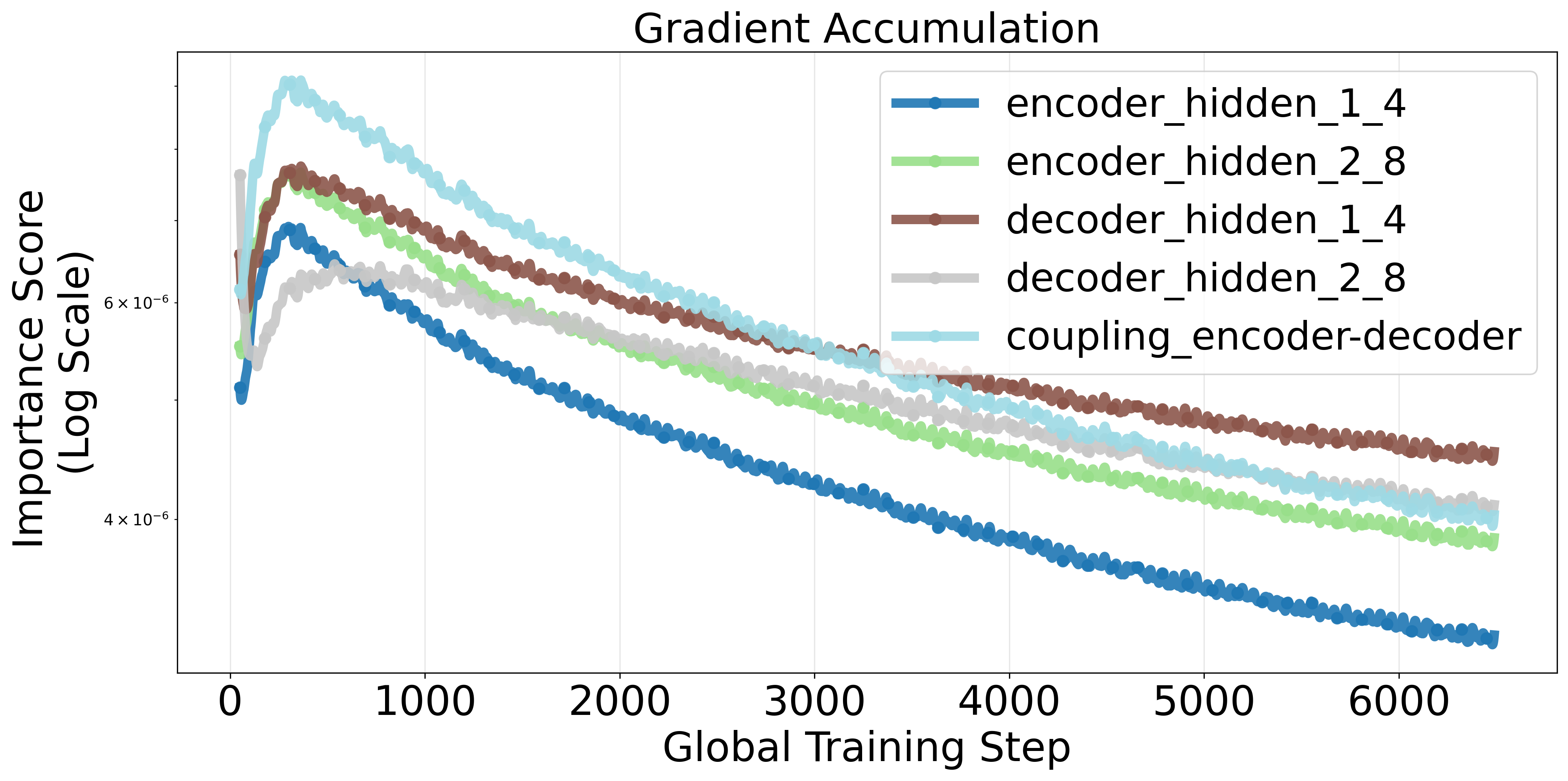} \hfill
        \includegraphics[width=0.32\linewidth]{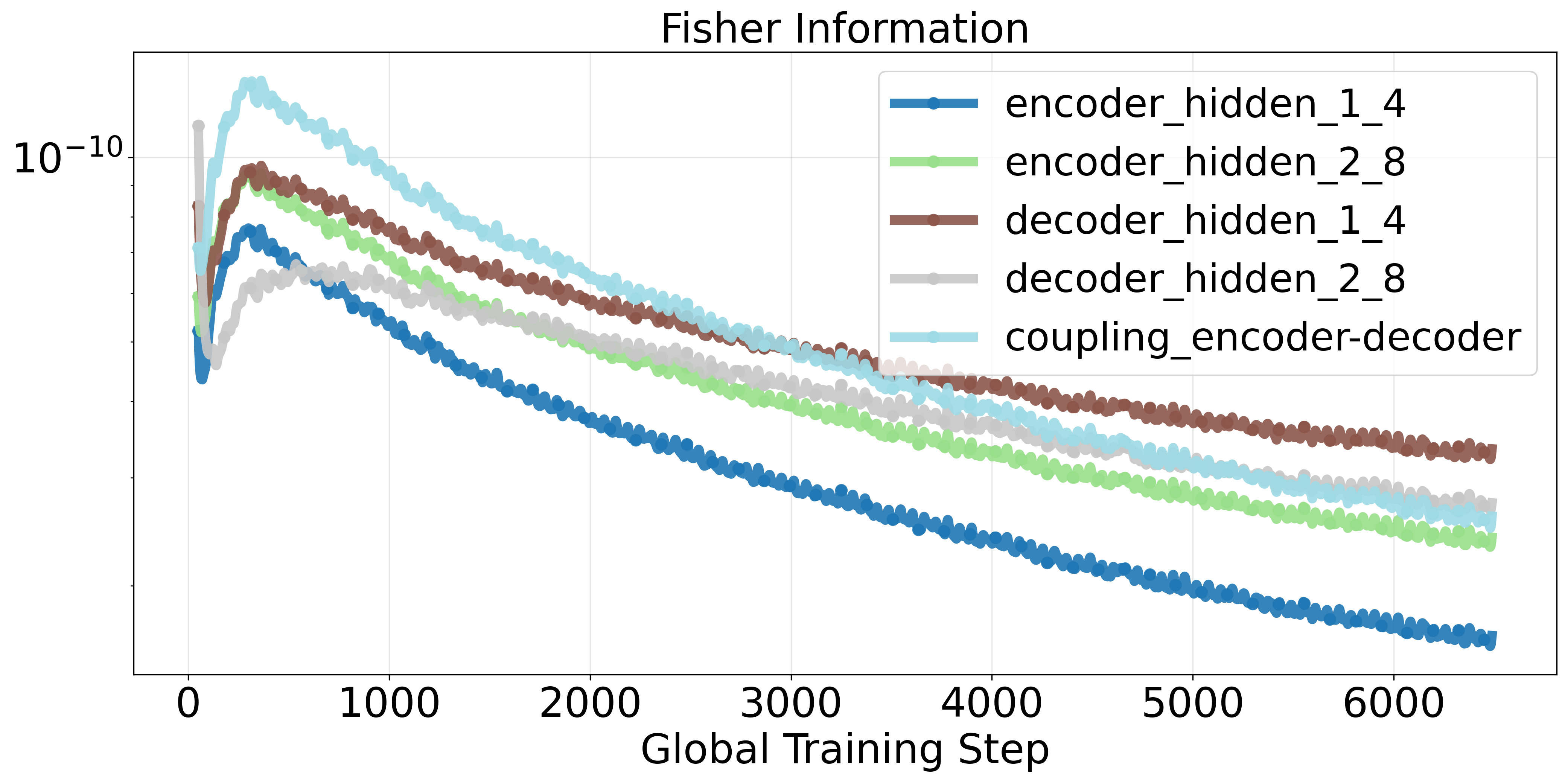} \hfill
        \includegraphics[width=0.32\linewidth]{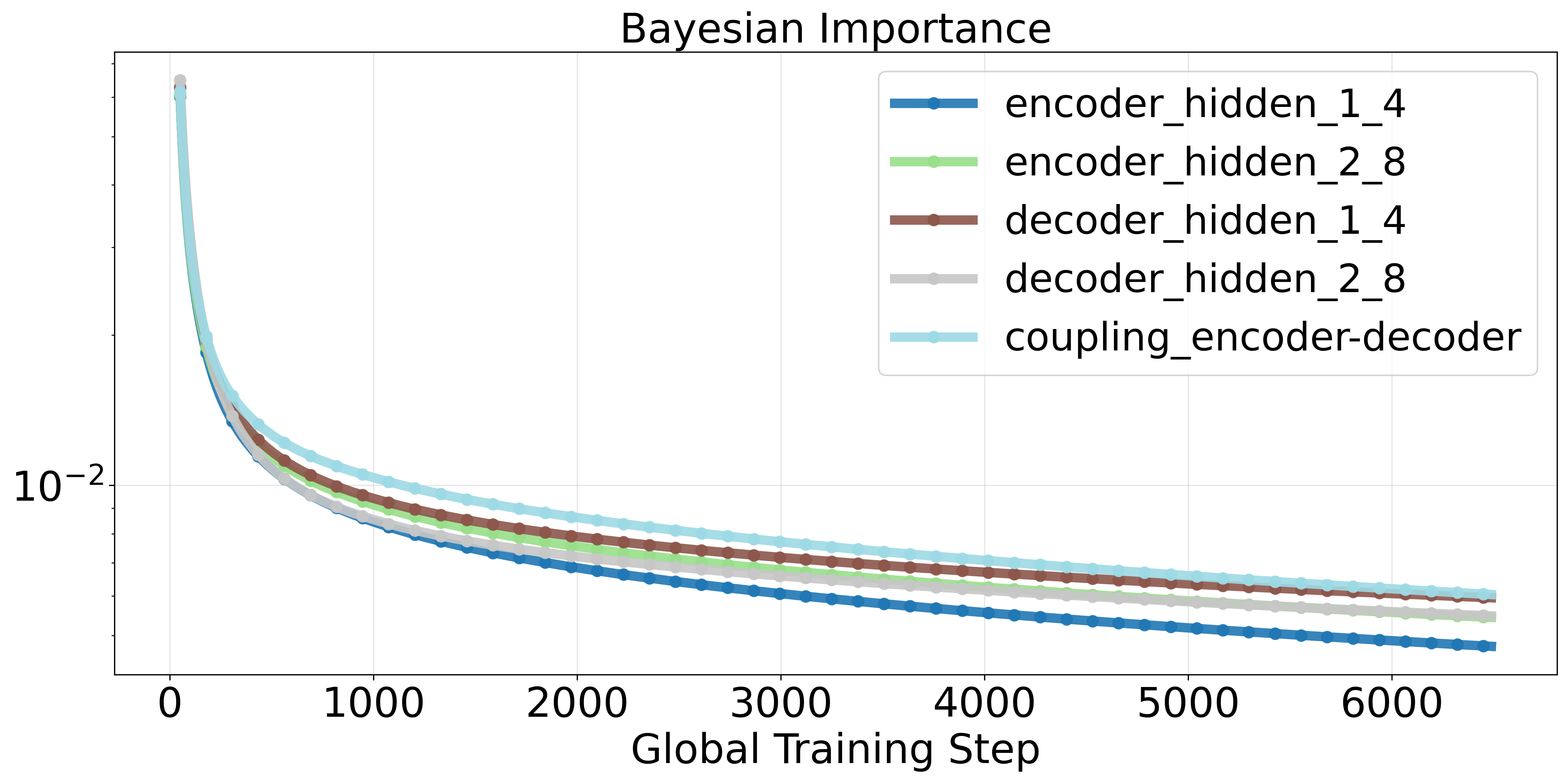}
        \caption{Latent Dimension 512 (High Capacity)}
        \label{fig:row4}
    \end{subfigure}

    \caption{Group importance scores compared across various Autoencoder architectures with latent dimension sizes ($d \in \{8, 64, 256, 512\}$) after 110 training epochs. Columns indicate estimation methods (Gradient, Fisher, Bayesian), while rows denote different architectures.}
    \label{fig:autoencoder_importance}
\end{figure*}

In Figure~\ref{fig:autoencoder_importance}, illustrating the evolution of importance scores across varying latent dimensions, it can be observed that the validity of \emph{Hypothesis 1}, which claims that coupling groups are inherently the most critical due to their role in component integration, is conditional rather than absolute. In tightly constrained architectures (e.g., $d \in \{8, 64\}$), the coupling group consistently demonstrates the highest importance scores across all three metrics. In this context, the coupling layer functions as a severe information bottleneck, so any pruning at this junction results in immediate and substantial information loss as the latent dimension increases to 512 (Figure~\ref{fig:autoencoder_importance}d). The importance of the coupling group decreases relative to component-specific groups, particularly the decoder layers. This crossover indicates that, in high-capacity models, the network depends less on the bottleneck for stability and more on internal processing. Therefore, coupling groups are not exempt from pruning in over-parameterized regimes.

\emph{Hypothesis 2} proposed that groups located early in the processing pipeline (e.g., the first encoder layer) are essential for preserving downstream functionality. The empirical data strongly refute this assumption. Across all configurations in Figure~\ref{fig:autoencoder_importance}, the first encoder group (\texttt{encoder\_hidden\_1\_4}) consistently exhibits the lowest importance scores. In contrast, the first decoder group (\texttt{decoder\_hidden\_1\_4}) and the second encoder group are often more sensitive. This disparity suggests that redundancy and relative size, rather than topological position, are more influential in determining a group's importance. The initial encoder layer, being large and over-parameterized for simple MNIST features, exhibits significant redundancy, allowing other groups to compensate for pruning. Conversely, the decoder layers are more critical for the final reconstruction task (MSE loss), making them more important regardless of their later position in the pipeline.

\begin{figure*}[ht]
    \centering
    \begin{subfigure}{\textwidth}
        \centering
        \includegraphics[width=0.32\linewidth]{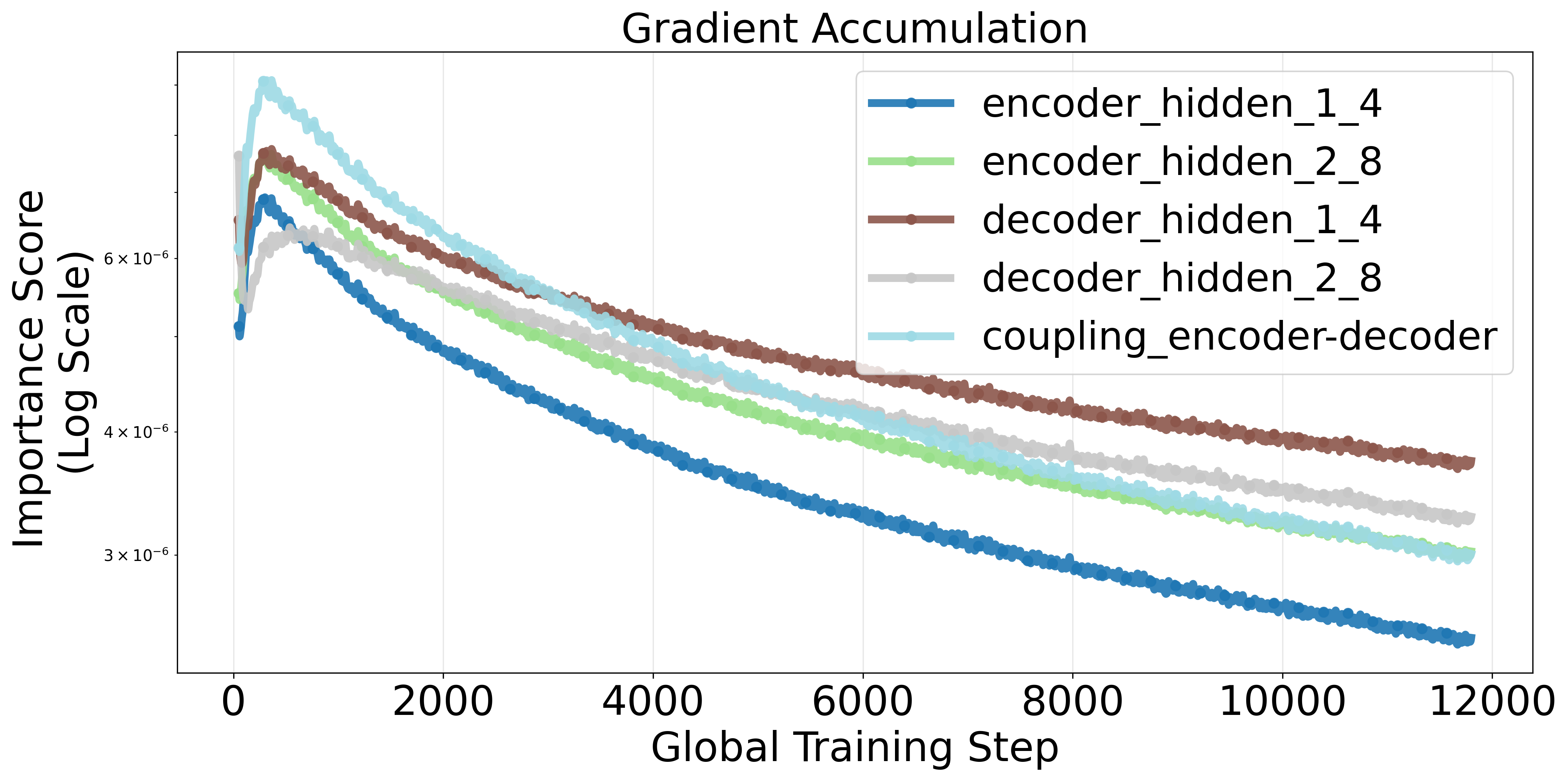} \hfill
        \includegraphics[width=0.32\linewidth]{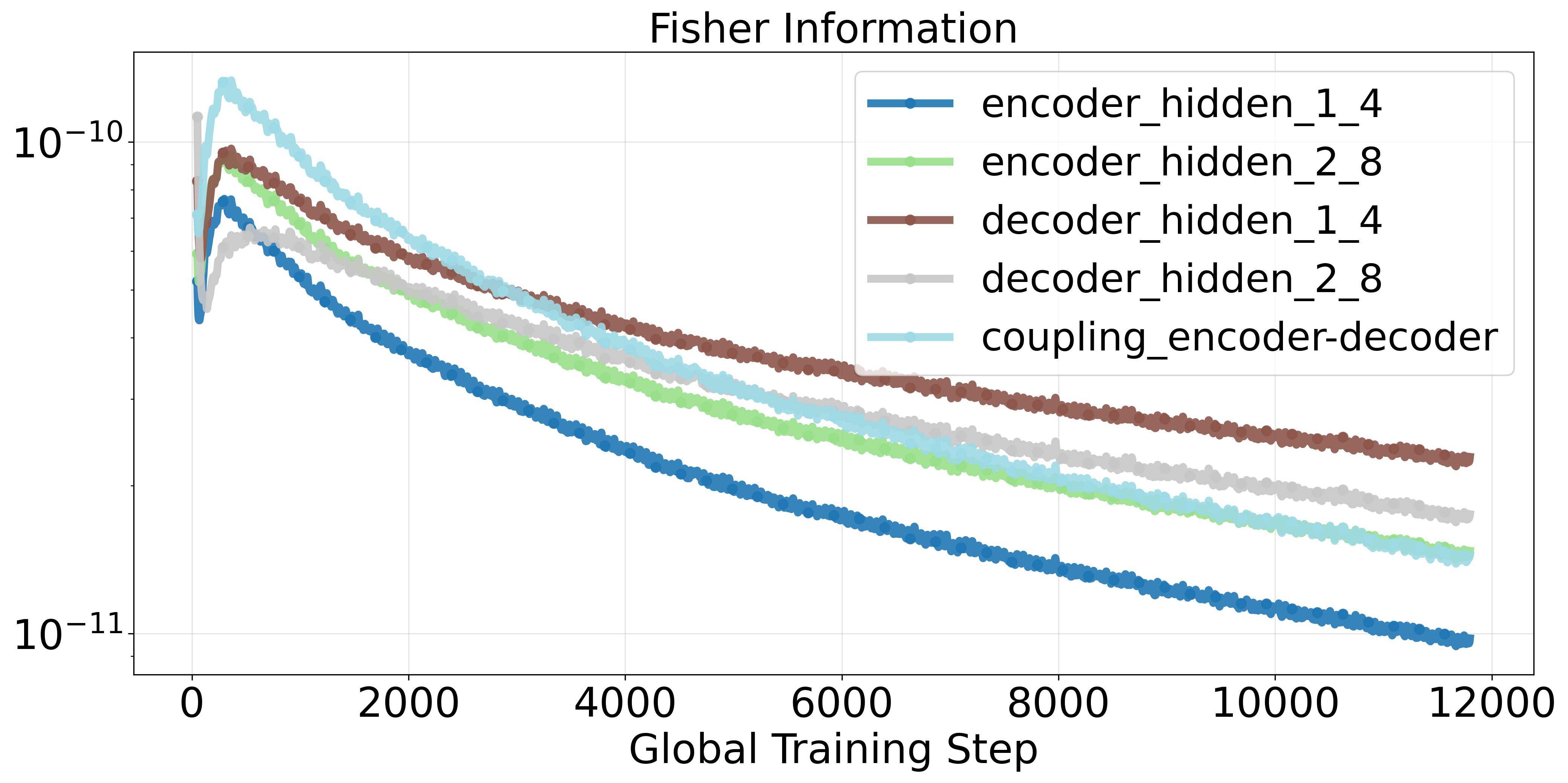} \hfill
        \includegraphics[width=0.32\linewidth]{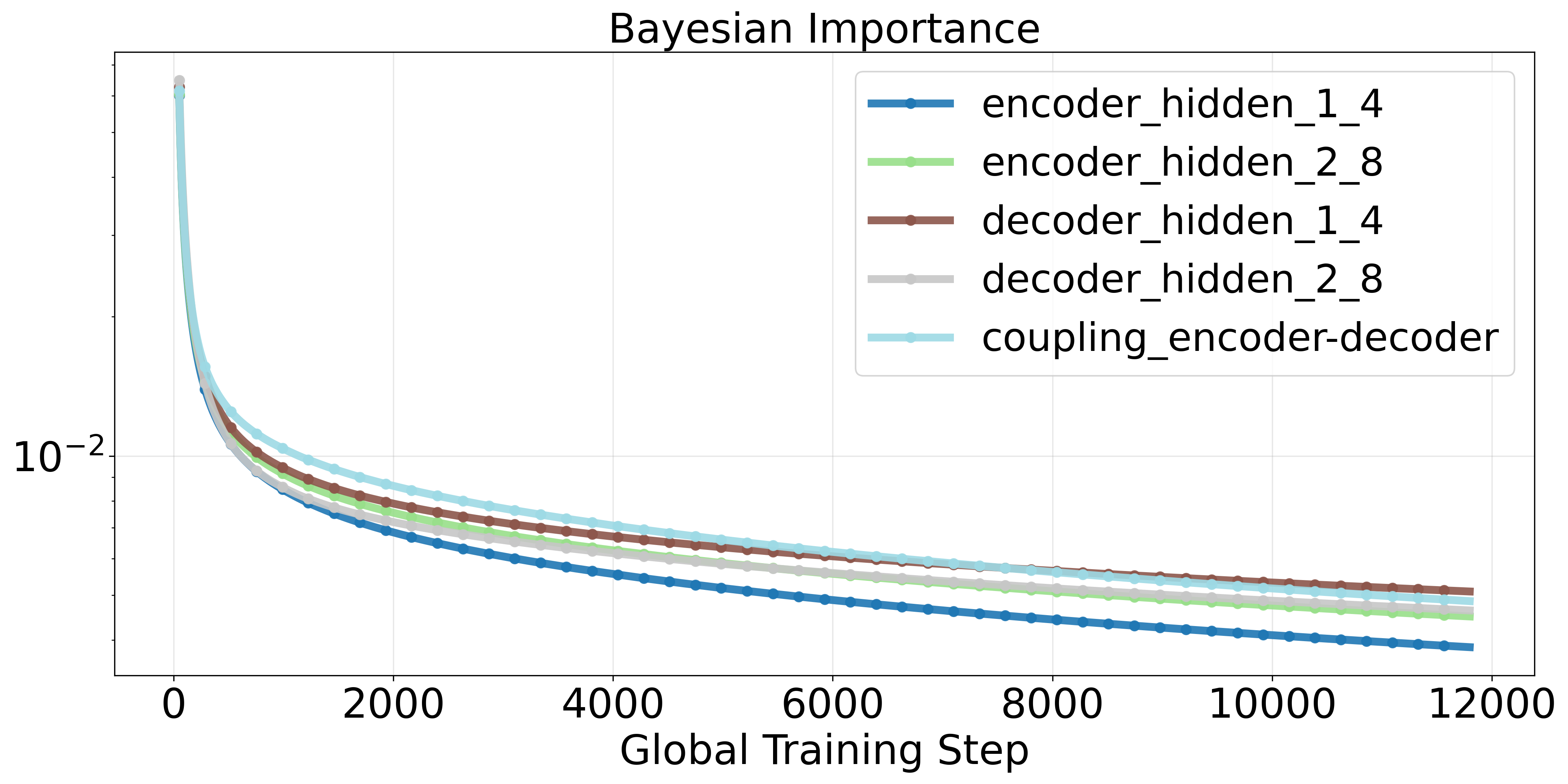}
    \end{subfigure}
    \caption{Extended training dynamics over 200 epochs for the 512-dimensional model. The observed crossovers in importance rankings, especially in the Gradient and Fisher metrics, challenge the static importance assumption stated in \emph{Hypothesis 3}.}
    \label{fig:ae_trained_200}
\end{figure*}

\emph{Hypothesis 3} assumed that group importance remains static throughout training. This assumption appears valid for highly constrained models, where the optimization landscape is narrow; however, it does not hold for models with higher capacity. Figure~\ref{fig:ae_trained_200} presents the extended training dynamics (200 epochs) for the 512-dimensional model. Distinct crossovers in the importance rankings are observed during the early-to-mid training phases, particularly in the Gradient and Fisher metrics. For example, the coupling group initially exhibits high importance but gradually yields dominance to the decoder layers as the model converges. These findings indicate that as network parameters adapt, the flow of critical information shifts. In the early stages, connectivity (coupling) is paramount, whereas in later stages, feature refinement (decoding) becomes the primary factor in minimizing loss. This demonstrates that importance is a dynamic property that evolves during the learning process.

\subsection{Use Case 2: TD-MPC for Balancing an Inverted Pendulum}

\begin{figure*}[ht]
    \centering
    \begin{subfigure}{\textwidth}
        \centering
        \includegraphics[width=0.49\linewidth]{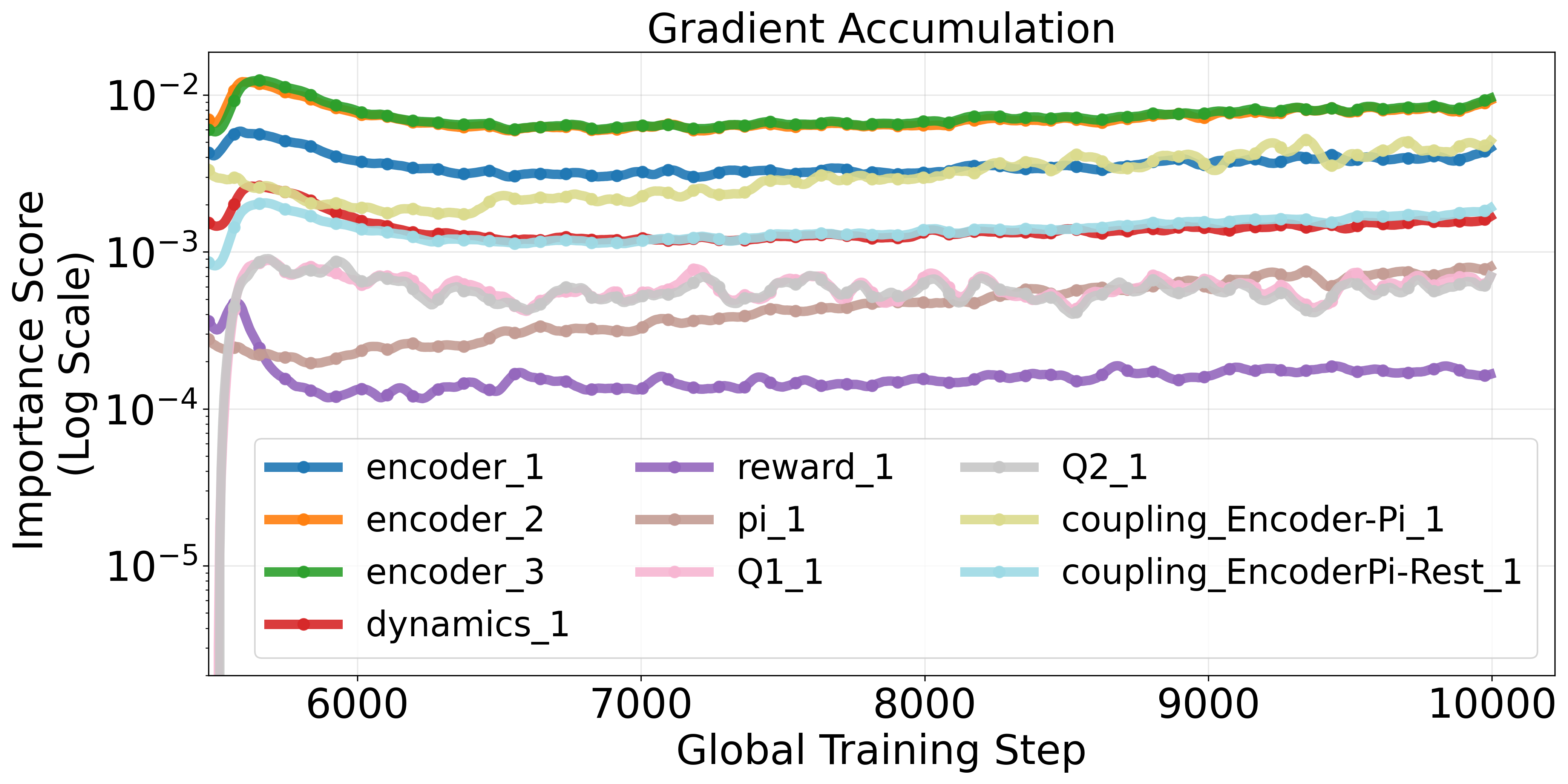} \hfill
        \includegraphics[width=0.49\linewidth]{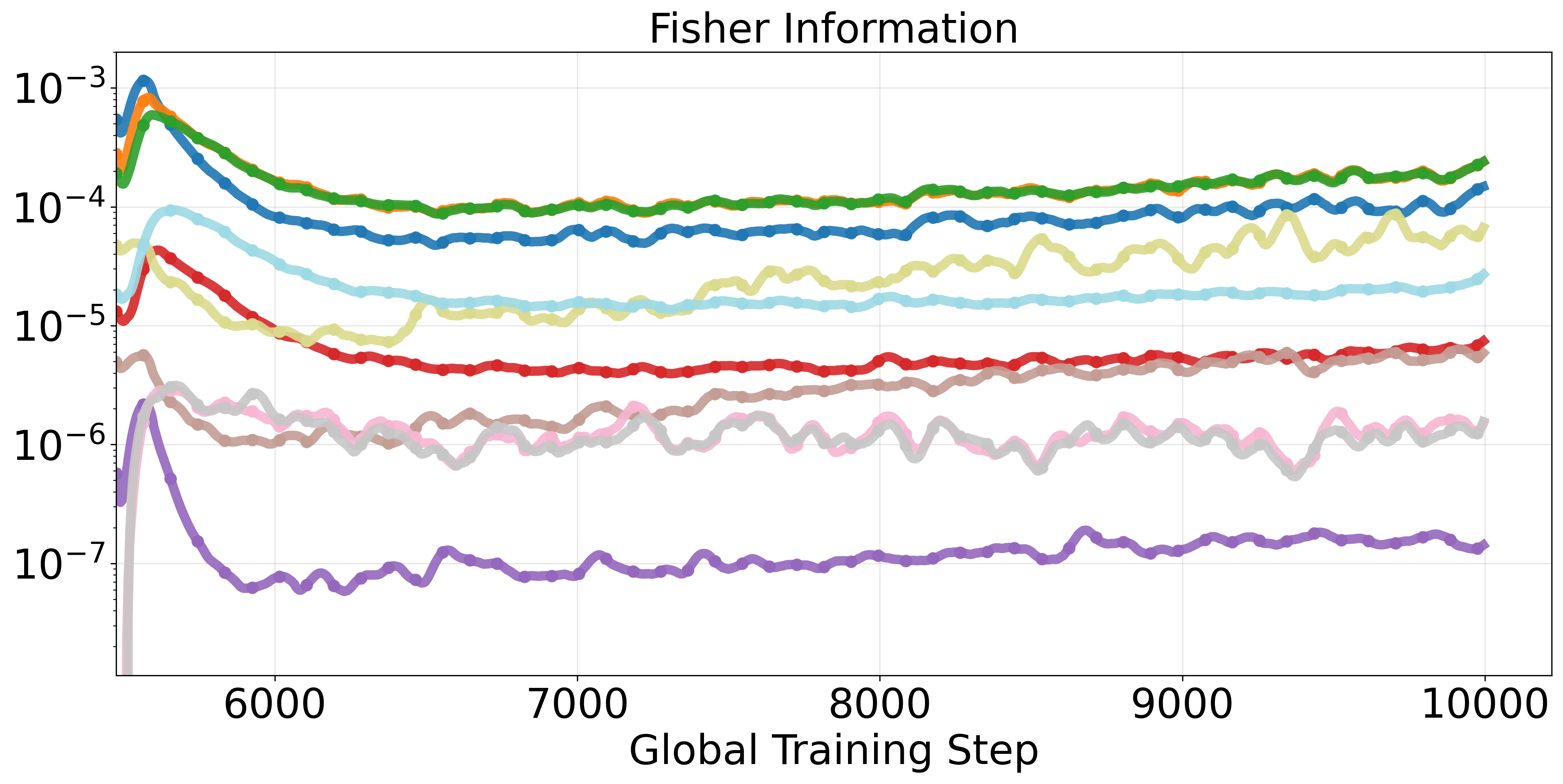}
    \end{subfigure}
    \caption{Evolution of group importance for TD-MPC using Gradient Accumulation (left) and Fisher Information (right). Encoder groups (green/orange) consistently exhibit greater importance than coupling layers, which contradicts \emph{Hypothesis 1}.}
    \label{fig:tdmpc_trained_gradient_fisher}
\end{figure*}

\begin{figure}[ht]
    \centering
    \includegraphics[width=\linewidth]{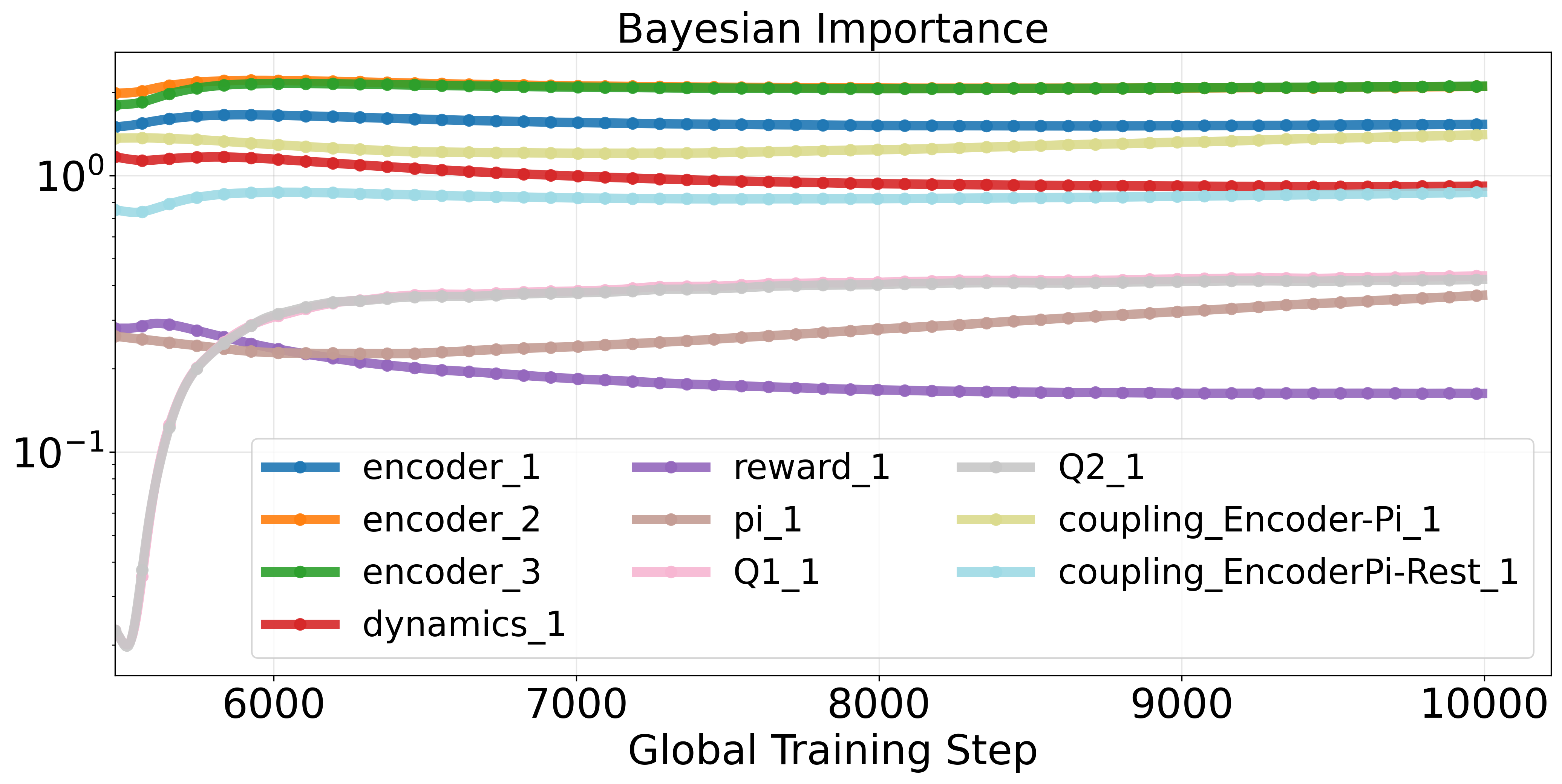}
    \caption{Bayesian importance estimates for TD-MPC. The pronounced decay of the reward model's importance (purple line) indicates that group criticality is dynamic and changes throughout training.}
    \label{fig:tdmpc_trained_bayesian}
\end{figure}

In the TDMPC controller, the MCNA architecture comprises an encoder with three groups (\texttt{encoder\_\{1,2,3\}}), a learned dynamics model (\texttt{dynamics\_1}), a reward model (\texttt{reward\_1}), a policy model (\texttt{pi\_1}), two value models (\texttt{Q1\_1}, \texttt{Q2\_1}), and two coupling groups. The evolution of their group importance during training is illustrated in Figure~\ref{fig:tdmpc_trained_gradient_fisher} and Figure~\ref{fig:tdmpc_trained_bayesian}. 

Similar to the previous use case, \emph{Hypothesis 1} is contradicted here as well. As shown in Figure~\ref{fig:tdmpc_trained_gradient_fisher}, both Gradient Accumulation (left) and Fisher Information (right) assign the highest importance to the encoder groups: \texttt{encoder\_2} and \texttt{encoder\_3} consistently rank above \texttt{encoder\_1} and \texttt{dynamics\_1}. The coupling and policy/value groups form a secondary tier, while \texttt{reward\_1} is consistently the least important. Therefore, in TDMPC, accurate representation learning and dynamics prediction are more essential for control performance than the interface layers that connect the encoder to the policy and value components.

\emph{Hypothesis 2}, which asserts that groups from early components are inherently more important, receives only partial support. While the encoder component is the most significant overall, within the encoder stack, the latter group \texttt{encoder\_3} is generally more critical than \texttt{encoder\_1}. This suggests that higher-level abstract features near the encoder output contribute more to performance than the raw features generated by the initial layer. The consistently low importance of the reward model further demonstrates that proximity to the input, or its early position in the computational graph, does not guarantee criticality.

Finally, \emph{Hypothesis 3} assumes that group importance remains static throughout training. In TDMPC, the relative ranking of the main groups stabilizes after an initial transient period; however, the importance curves continue to display temporal dynamics. Figure~\ref{fig:tdmpc_trained_bayesian} demonstrates that the Bayesian importance of \texttt{reward\_1} declines steadily over time. Similarly, the Gradient and Fisher scores for the coupling and dynamics groups exhibit gradual changes rather than remaining constant. Thus, group importance is not strictly static but adapts as the controller refines its policy and value estimates.

\section{Conclusion and Future Work}

This study introduces a component-aware structured pruning framework that addresses the limitations of static, heuristic-based pruning in MCNAs. Instead of relying solely on weight norms, the proposed method utilizes three gradient-driven importance measures: Gradient Accumulation, Fisher Information, and Bayesian Uncertainty. These measures elucidate the contributions of parameter groups to task performance and guide pruning decisions to achieve targeted sparsity. Experimental evaluations of an autoencoder and a TD-MPC controller show that explicitly distinguishing between component-specific and coupling groups improves both pruning flexibility and accuracy. Furthermore, the findings reveal that group importance is not static; it depends on the architecture and evolves dynamically during training, thereby challenging the assumption that layers or couplings are universally critical.

A key contribution of this work is the analysis of how importance metrics evolve during training. Because the framework continuously assesses importance, it can inform pruning decisions at any stage of training. This enables flexible workflows, such as pruning the model at convergence for immediate deployment (Train $\to$ Prune) or pruning during training and continuing optimization with the reduced model (Train $\to$ Prune $\to$ Continue Training). By using instantaneous (Gradient/Fisher) or historical (Bayesian) metrics, the approach enables informed, data-driven pruning decisions without the substantial computational cost of post-training sensitivity analysis.

Future research will extend this framework to formally investigate the effects of intermediate pruning steps on final model performance. Additionally, theoretical stability guarantees from a control perspective, such as integrating Lyapunov stability criteria and formal sensitivity analysis into the importance estimation process, will be incorporated. These enhancements are crucial for deploying pruned NNCs in safety-critical embedded systems, ensuring that model compression does not compromise the robustness or safety of the underlying control policy.

\bibliography{ifacconf}             

\end{document}

%% file: introduction.tex
The adoption of Deep Neural Networks (DNNs) has transformed numerous domains, including perception and control, by facilitating advanced representation learning and generalization. Despite these advances, the increasing complexity of multi-component model architectures for environmental understanding has widened the gap between research-grade models and the strict resource limitations of real-world embedded systems~\citep{liu2025survey}. Closing this gap necessitates effective model compression. However, conventional approaches such as unstructured pruning or magnitude-based structured pruning often treat networks as indivisible entities. Traditional methods typically use static, rule-based heuristics that fail to account for cross-component dependencies. This oversight can result in the elimination of critical connections that are essential for maintaining the stability and performance of complex closed-loop controllers.

This work addresses these limitations by introducing a component-aware structured pruning framework. Unlike conventional methods, the proposed approach explicitly distinguishes between component-specific groups and inter-component coupling groups, enabling a more granular and adaptable pruning strategy. Three prevailing heuristics for assessing group importance are systematically evaluated using two representative scenarios: a canonical Multi-Component Neural Architecture (MCNA) autoencoder and a control-oriented Temporal Difference Learning for Model Predictive Control (TD-MPC) agent. The empirical results challenge the prevailing assumption that coupling groups or early layers are inherently critical. Instead, the findings indicate that group importance is conditional, architecture-dependent, and highly dynamic, with significant shifts occurring throughout training. To capture these dynamics, the framework incorporates a unified pruning pipeline that assesses importance using three distinct metrics: \emph{Gradient Accumulation}, \emph{Fisher Information}, and \emph{Bayesian Uncertainty}. This allows for informed, data-driven pruning decisions without incurring the high computational cost of post-training sensitivity analysis. By revealing essential structural dependencies, the method preserves control performance under aggressive compression and offers interpretable insights into the contributions of different components to overall controller fidelity.

The remainder of this paper is organized as follows: Section~\ref{sec:related_work} reviews prior pruning methods, Section~\ref{sec:methodology} details the structured pruning formalism and importance measures, Section~\ref{sec:experimental_step} describes the experimental setup, and Section~\ref{sec:results_discussion} presents the results and analysis.

%% file: related_works/related_work_haiku.tex
Structured pruning methods are typically classified into magnitude-based heuristics, importance estimation techniques, and learning-based optimization approaches.

\subsection{Magnitude-Based Criteria}

Early heuristic methods performed pruning based on activation patterns or magnitude ranking, sometimes employing an \(\ell_1\) penalty to promote sparsity \citep{hu_network_2016,li_pruning_nodate,liu_learning_2017}. Structured sparsity can be applied to channels, filters, or blocks and may be guided by reconstruction error or gating functions~\citep{luo_thinet_2017,he_channel_2017,he_channel_2017-1}. Gradually increasing sparsity regularization has been shown to reduce redundancy~\citep{wang_structured_2020}. However, recent surveys caution that naive heuristics risk discarding essential parameters and emphasize that training the target architecture from scratch can achieve performance comparable to pruned models~\citep{he_structured_2024,cheng_survey_2024,blalock_what_2020}.

\subsection{Importance Estimation and Second-Order Methods}

Second-order methods, including Optimal Brain Damage, Optimal Brain Surgeon, and Taylor-series pruning, approximate the Hessian matrix to evaluate parameter importance~\citep{lecun_optimal_1989,frantar_optimal_2023,molchanov_pruning_2017}. More recent approaches use gradient accumulation and Bayesian uncertainty to estimate the importance of parameter groups~\citep {molchanov_importance_2019,molchanov_importance_2019-1,ke_optimization_2022}.

\subsection{Optimization Frameworks}

Learning-based approaches conceptualize pruning as a search or optimization problem. Reinforcement learning and automated machine learning (AutoML) methods are employed to explore optimal sparsity ratios~\citep{huang_learning_2018}. Single-shot saliency techniques, such as SNIP~\citep {lee_snip_2019}, assess parameter importance at initialization. Hybrid frameworks enable the simultaneous pruning of layers, attention heads, and hidden units in transformer architectures~\citep{michel_are_2019,xia_structured_2022}. Collectively, these methods demonstrate that pruning can be integrated within a comprehensive optimization pipeline rather than relying solely on fixed heuristics.

%% file: ifacconf.bbl
\begin{thebibliography}{27}
\providecommand{\natexlab}[1]{#1}
\providecommand{\url}[1]{\texttt{#1}}
\providecommand{\urlprefix}{URL }
\expandafter\ifx\csname urlstyle\endcsname\relax
  \providecommand{\doi}[1]{doi:\discretionary{}{}{}#1}\else
  \providecommand{\doi}{doi:\discretionary{}{}{}\begingroup \urlstyle{rm}\Url}\fi

\bibitem[{Blalock et~al.(2020)Blalock, Ortiz, Frankle, and Guttag}]{blalock_what_2020}
Blalock, D., Ortiz, J.J.G., Frankle, J., and Guttag, J. (2020).
\newblock {What is the State of Neural Network Pruning?}
\newblock \emph{{arXiv:2003.03033}}.

\bibitem[{Brockman et~al.(2016)Brockman, Cheung, Pettersson, Schneider, Schulman, Tang, and Zaremba}]{brockman2016openai}
Brockman, G., Cheung, V., Pettersson, L., Schneider, J., Schulman, J., Tang, J., and Zaremba, W. (2016).
\newblock {OpenAI Gym}.
\newblock \emph{{arXiv:1606.01540}}.

\bibitem[{Cheng et~al.(2024)Cheng, Zhang, and Shi}]{cheng_survey_2024}
Cheng, H., Zhang, M., and Shi, J.Q. (2024).
\newblock {A Survey on Deep Neural Network Pruning: Taxonomy, Comparison, Analysis, and Recommendations}.
\newblock \emph{{IEEE Transactions on Pattern Analysis and Machine Intelligence}}, 46(12), 10558--10578.

\bibitem[{Fang et~al.(2023)Fang, Ma, Song, Mi, and Wang}]{Fang_2023_CVPR}
Fang, G., Ma, X., Song, M., Mi, M.B., and Wang, X. (2023).
\newblock {Depgraph: Towards Any Structural Pruning}.
\newblock In \emph{{Proceedings of the 2023 IEEE/CVF Conference on Computer Vision and Pattern Recognition}}, 16091--16101.

\bibitem[{Frantar et~al.(2023)Frantar, Singh, and Alistarh}]{frantar_optimal_2023}
Frantar, E., Singh, S.P., and Alistarh, D. (2023).
\newblock {Optimal Brain Compression: A Framework for Accurate Post-Training Quantization and Pruning}.
\newblock \emph{{arXiv:2208.11580}}.

\bibitem[{Hansen et~al.(2022)Hansen, Wang, and Su}]{hansen2022temporaldifferencelearningmodel}
Hansen, N., Wang, X., and Su, H. (2022).
\newblock {Temporal Difference Learning for Model Predictive Control}.
\newblock In \emph{{Proceedings of the 39th International Conference on Machine Learning}}, 8387--8406.

\bibitem[{He and Xiao(2024)}]{he_structured_2024}
He, Y. and Xiao, L. (2024).
\newblock {Structured Pruning for Deep Convolutional Neural Networks: A Survey}.
\newblock \emph{{IEEE Transactions on Pattern Analysis and Machine Intelligence}}, 46(5), 2900--2919.

\bibitem[{He et~al.(2017{\natexlab{a}})He, Zhang, and Sun}]{he_channel_2017}
He, Y., Zhang, X., and Sun, J. (2017{\natexlab{a}}).
\newblock {Channel Pruning for Accelerating Very Deep Neural Networks}.
\newblock In \emph{{Proceedings of the 2017 IEEE International Conference on Computer Vision (ICCV)}}, 1398--1406. IEEE.

\bibitem[{He et~al.(2017{\natexlab{b}})He, Zhang, and Sun}]{he_channel_2017-1}
He, Y., Zhang, X., and Sun, J. (2017{\natexlab{b}}).
\newblock {Channel Pruning for Accelerating Very Deep Neural Networks}.
\newblock In \emph{{Proceedings of the 2017 IEEE International Conference on Computer Vision (ICCV)}}, 1398--1406.
\newblock ISSN: 2380-7504.

\bibitem[{Hu et~al.(2016)Hu, Peng, Tai, and Tang}]{hu_network_2016}
Hu, H., Peng, R., Tai, Y.W., and Tang, C.K. (2016).
\newblock {Network Trimming: A Data-Driven Neuron Pruning Approach Towards Efficient Deep Architectures}.
\newblock \emph{{arXiv:1607.03250}}.

\bibitem[{Huang et~al.(2018)Huang, Zhou, You, and Neumann}]{huang_learning_2018}
Huang, Q., Zhou, K., You, S., and Neumann, U. (2018).
\newblock {Learning to Prune Filters in Convolutional Neural Networks}.
\newblock \emph{{arXiv:1801.07365}}.

\bibitem[{Ke and Fan(2022)}]{ke_optimization_2022}
Ke, X. and Fan, Y. (2022).
\newblock {On the Optimization and Pruning for Bayesian Deep Learning}.
\newblock \emph{{arXiv:2210.12957}}.

\bibitem[{LeCun et~al.(1989)LeCun, Denker, and Solla}]{lecun_optimal_1989}
LeCun, Y., Denker, J., and Solla, S. (1989).
\newblock {Optimal Brain Damage}.
\newblock In \emph{{Advances in Neural Information Processing Systems}}, volume~2. Morgan-Kaufmann.

\bibitem[{Lee et~al.(2019)Lee, Ajanthan, and Torr}]{lee_snip_2019}
Lee, N., Ajanthan, T., and Torr, P.H.S. (2019).
\newblock {SNIP: Single-Shot Network Pruning Based on Connection Sensitivity}.
\newblock \emph{{arXiv:1810.02340}}.

\bibitem[{Li et~al.(2016)Li, Kadav, Durdanovic, Samet, and Graf}]{li_pruning_nodate}
Li, H., Kadav, A., Durdanovic, I., Samet, H., and Graf, H. (2016).
\newblock {Pruning Filters for Efficient Convnets}.
\newblock \emph{{arXiv:1608.08710}}.

\bibitem[{Liu et~al.(2025)Liu, Zhu, Liu, Liu, Han, Tian, Li, and Yi}]{liu2025survey}
Liu, D., Zhu, Y., Liu, Z., Liu, Y., Han, C., Tian, J., Li, R., and Yi, W. (2025).
\newblock {A Survey of Model Compression Techniques: Past, Present, and Future}.
\newblock \emph{{Proceedings of the Frontiers in Robotics and AI}}, 12, 1518965.

\bibitem[{Liu et~al.(2017)Liu, Li, Shen, Huang, Yan, and Zhang}]{liu_learning_2017}
Liu, Z., Li, J., Shen, Z., Huang, G., Yan, S., and Zhang, C. (2017).
\newblock {Learning Efficient Convolutional Networks Through Network Slimming}.
\newblock \emph{{arXiv:1708.06519}}.

\bibitem[{Luo et~al.(2017)Luo, Wu, and Lin}]{luo_thinet_2017}
Luo, J.H., Wu, J., and Lin, W. (2017).
\newblock {Thinet: A Filter Level Pruning Method for Deep Neural Network Compression}.
\newblock In \emph{{In Proceedings of the 2017 IEEE International Conference on Computer Vision (ICCV)}}.

\bibitem[{Michel et~al.(2019)Michel, Levy, and Neubig}]{michel_are_2019}
Michel, P., Levy, O., and Neubig, G. (2019).
\newblock {Are Sixteen Heads Really Better Than One?}
\newblock In \emph{{Advances in Neural Information Processing Systems}}, volume~32. Curran Associates, Inc.

\bibitem[{Molchanov et~al.(2019{\natexlab{a}})Molchanov, Mallya, Tyree, Frosio, and Kautz}]{molchanov_importance_2019}
Molchanov, P., Mallya, A., Tyree, S., Frosio, I., and Kautz, J. (2019{\natexlab{a}}).
\newblock {Importance Estimation for Neural Network Pruning}.
\newblock In \emph{{Proceedings of the 2019 IEEE/CVF Conference on Computer Vision and Pattern Recognition (CVPR)}}, 11256--11264.

\bibitem[{Molchanov et~al.(2019{\natexlab{b}})Molchanov, Mallya, Tyree, Frosio, and Kautz}]{molchanov_importance_2019-1}
Molchanov, P., Mallya, A., Tyree, S., Frosio, I., and Kautz, J. (2019{\natexlab{b}}).
\newblock {Importance Estimation for Neural Network Pruning}.
\newblock In \emph{{Proceedings of the IEEE/CVF Conference on Computer Vision and Pattern Recognition}}, 11264--11272.

\bibitem[{Molchanov et~al.(2017)Molchanov, Tyree, Karras, Aila, and Kautz}]{molchanov_pruning_2017}
Molchanov, P., Tyree, S., Karras, T., Aila, T., and Kautz, J. (2017).
\newblock {Pruning Convolutional Neural Networks for Resource Efficient Inference}.
\newblock \emph{{arXiv:1611.06440}}.

\bibitem[{Sundaram et~al.(2025{\natexlab{a}})Sundaram, Ulmen, and G\"orges}]{sundaram2025enhancedpruningstrategymulticomponent}
Sundaram, G., Ulmen, J., and G\"orges, D. (2025{\natexlab{a}}).
\newblock {Enhanced Pruning Strategy for Multi-Component Neural Architectures Using Component-Aware Graph Analysis}.
\newblock In \emph{{Proceedings of the IFAC Joint Conference on Computers, Cognition and Communication (J3C)}}.

\bibitem[{Sundaram et~al.(2025{\natexlab{b}})Sundaram, Ulmen, Haider, and G{\"o}rges}]{sundaram2025application}
Sundaram, G., Ulmen, J., Haider, A., and G{\"o}rges, D. (2025{\natexlab{b}}).
\newblock {Application-Specific Component-Aware Structured Pruning of Deep Neural Networks via Soft Coefficient Optimization}.
\newblock \emph{{arXiv:2507.14882}}.

\bibitem[{Wang et~al.(2020)Wang, Hu, Zhang, Wang, Yu, and Hu}]{wang_structured_2020}
Wang, H., Hu, X., Zhang, Q., Wang, Y., Yu, L., and Hu, H. (2020).
\newblock {Structured Pruning for Efficient Convolutional Neural Networks via Incremental Regularization}.
\newblock \emph{{IEEE Journal of Selected Topics in Signal Processing}}, 14(4), 775--788.

\bibitem[{Xia et~al.(2022)Xia, Zhong, and Chen}]{xia_structured_2022}
Xia, M., Zhong, Z., and Chen, D. (2022).
\newblock {Structured Pruning Learns Compact and Accurate Models}.
\newblock In S.~Muresan, P.~Nakov, and A.~Villavicencio (eds.), \emph{{Proceedings of the 60th Annual Meeting of the Association for Computational Linguistics (Volume 1: Long Papers)}}, 1513--1528. Association for Computational Linguistics.

\bibitem[{Zhu(2018)}]{zhu2018classification}
Zhu, W. (2018).
\newblock {Classification of MNIST Handwritten Digit Database Using Neural Network}.
\newblock In \emph{{Proceedings of the Research School of Computer Science. Australian National University, Acton, ACT}}, volume 2601.

\end{thebibliography}
